\documentclass[sigconf,natbib=false]{acmart}

\usepackage{graphicx}
\usepackage{algorithmic}
\usepackage{subcaption}
\usepackage{tabularx}
\usepackage{tikz}
\usepackage{tikz-cd}
\usetikzlibrary{shapes,arrows,fit,positioning,backgrounds,calc}
\usepackage{pgfplots}
\pgfplotsset{compat=1.18}
\usepackage{stackengine,scalerel}
\usepackage{todonotes}
\usepackage{mathtools}
\usepackage{cancel}
\usepackage{svg} 
\usepackage{adjustbox}
\usepackage{array}
\usepackage{booktabs}
\usepackage{multirow}
\usepackage{pifont}
\usepackage{comment}
\usepackage{mfirstuc}
\usepackage{chemfig}
\usepackage{arydshln}
\usepackage{enumitem}
\usepackage{tcolorbox}
\usepackage{fancyhdr}
\usepackage{hyperref}

\AtBeginDocument{%
  }

\RequirePackage[
  datamodel=acmdatamodel,
  style=acmnumeric,
  ]{biblatex}

\copyrightyear{2026}
\acmYear{2026}
\setcopyright{rightsretained} 
\acmConference[SAC '26]{The 41st ACM/SIGAPP Symposium on Applied Computing}{March 23--27, 2026}{Thessaloniki, Greece}
\acmBooktitle{The 41st ACM/SIGAPP Symposium on Applied Computing (SAC '26), March 23--27, 2026, Thessaloniki, Greece}
\acmPrice{}
\acmDOI{10.1145/3748522.3779865}
\acmISBN{979-8-4007-2294-3/2026/03}

\begin{document}
\definecolor{CornflowerBlue}{HTML}{6495ED}

\title{A Subjective Logic-based method for runtime confidence updates in safety arguments}
\renewcommand{\shorttitle}{Integrating Runtime Evidence Into SL-Based Assurance Arguments}
\renewcommand{\shorttitle}{An SL-based method for runtime confidence updates in safety arguments}

\author{Benjamin Herd}
\email{benjamin.herd@iks.fraunhofer.de}
\orcid{0000-0001-6439-8845}
\affiliation{%
  \institution{Fraunhofer Institute for Cognitive Systems (IKS)}
  \city{Munich}
  \country{Germany}
}

\author{Jessica Kelly}
\email{jessica.kelly@iks.fraunhofer.de}
\orcid{0009-0003-0508-2367}
\affiliation{%
  \institution{Fraunhofer Institute for Cognitive Systems (IKS)}
  \city{Munich}
  \country{Germany}
}

\author{Clarissa Heinemann}
\email{clarissa.heinemann@gmail.com}
\orcid{0009-0004-4974-1608}
\affiliation{%
  \institution{Fraunhofer Institute for Cognitive Systems (IKS)}
  \city{Munich}
  \country{Germany}
}

\author{Jo\~{a}o-Vitor Zacchi}
\email{joao-vitor.zacchi@iks.fraunhofer.de}
\orcid{0009-0002-6975-6829}
\affiliation{%
  \institution{Fraunhofer Institute for Cognitive Systems (IKS)}
  \city{Munich}
  \country{Germany}
}


\begin{abstract}
  We present a method for dynamic quantitative assurance that enhances static safety cases with continuous, runtime-driven confidence updates. The method quantifies and propagates confidence across the development lifecycle by integrating design-time evidence and windowed runtime Safety Performance Indicators (SPIs) within a single Subjective Logic (SL)-based assurance case. At runtime, SPI evidence is continuously evaluated, and targeted claims are updated using a rule that increases confidence in the absence of violations and imposes prompt penalties when violations occur. This design prioritizes safety-relevant responsiveness over exact classical Bayesian posterior updates. We demonstrate the method using a simulation-based construction zone assist function, focusing on an ML-based construction cone detection component, and show how confidence evolves as SPI evidence is observed in operation.
\end{abstract}



\maketitle


\section{Introduction}
Safety assurance arguments for machine learning (ML) components offer structured justifications that a system is safe to operate within a defined operational context. As these systems are deployed in complex and dynamic environments, there is a need to reflect the evolving confidence in the \textit{validity} of the assurance argument over time. According to UL 4600 \cite{UL4600_2023}, an argument is valid if (1) all claims are supported by evidence, and (2) the evidence itself is valid, i.e.\ supported by factual, objective data. However, given the finite amount of training and testing data and the incomplete understanding of the operational context including potential triggering conditions, there is always a chance of the argument becoming invalidated over time. Shifts in data distribution can lead to model drift and challenge the continuing relevance of the initial argument; assumptions about the surrounding system and the operational context may be invalidated by evolving requirements or the integration of new technologies. It is thus crucial to adopt a systematic approach for ongoing reassessment of assurance arguments and to ensure sustained confidence and alignment with operational demands. 

One way of incorporating runtime information into assurance arguments is through \textit{Safety Performance Indicators (SPIs)}. SPIs are attached to claims in the argument and defined as ``metrics supported by evidence that use threshold comparisons of condition claims in a safety case'' \cite{UL4600_2023}. The metric values are empirically gathered by observing the system throughout its lifecycle. 

In this paper, we extend prior work \cite{herd2024} \cite{herd2025} on the formulation of static, quantitative assurance arguments using Subjective Logic (SL) by incorporating runtime information for the purpose of \textit{continuous assurance}. More precisely, we integrate the concept of SPIs into quantitative SL-based assurance arguments to study the evolution of confidence in the assurance case, enabling a more dynamic and responsive approach to safety validation. The approach allows for runtime evidence to target specific claims and propagate through the same formalism (SL) used for computing assurance confidence in the static argument. We deliberately go beyond classical probabilistic or Bayesian update by proposing an update rule that increases confidence when no SPI violations occur and incurs an immediate penalty in the case of violations. The approach thus prioritizes safety-relevant responsiveness over exact conjugate-posterior equivalence in a purely Bayesian setting. 

Our contribution is an SL-native, explicitly non-Bayesian update method that integrates windowed runtime SPIs into quantitative assurance by choosing SL operators which enable gradual confidence accumulation in the absence of violations and immediate penalties when violations occur. We demonstrate the method using an ML-based construction cone detection simulation, illustrating the evolution of first- and second-order confidence in the argument during operational use. 

 
The paper is structured as follows: Section \ref{sec:background} provides background information on assurance confidence estimation, SPIs, and SL. Section \ref{sec:methodology} introduces an SL-based method for modelling confidence propagation from SPIs to claims in dynamic assurance arguments and illustrates it with a simulation of ML-based construction cone detection. Related work is presented in Section \ref{sec:related_work}. The paper concludes with a discussion in Section \ref{sec:conclusions}. 

\section{Background}\label{sec:background}

\subsection{Assurance confidence estimation}\label{sec:assurance_confidence}

According to ISO/IEC 15026, an assurance argument is ``a reasoned, auditable artifact that supports the contention that its top-level claim is satisfied, including systematic arguments and its underlying evidence and explicit assumptions that support the claim(s)'' \cite{IEEE_15026}. Assurance arguments are used to demonstrate that safety-critical systems satisfy their safety requirements in a given context. 

Assurance confidence assessments determine how much trust can be placed in a given argument. Various confidence assessment methods have been proposed and can be broadly classified into qualitative and quantitative approaches. Qualitative methods typically rely on expert judgment and involve dialectical techniques such as \textit{defeaters}, i.e.\ attacks that question the validity of specific claims or evidence within the argument. Defeaters play a key role in identifying potential weaknesses, gaps in reasoning, or uncertainties that could compromise the overall strength of safety assurance. Qualitative defeaters exert a binary effect on their target: they either refute it or leave it unchanged. On the other hand, quantitative methods for assurance confidence estimation aim to deliver numerical valuations of confidence through the use of mathematical and statistical approaches, e.g.\ Baconian probabilities \cite{Goodenough2013,graydonDefiningBaconianProbability2016}, Bayesian inference \cite{Guo2003,Denney2011,Hobbs2012}, Dempster-Shafer belief functions \cite{Ayoub2013,Wang2019}, or SL \cite{duan2016representation,yuan2017subjective,herd2024}. As discussed below, the nature of confidence assessment influences the representation of SPIs and their impact on the underlying claims.

\subsection{Safety Performance Indicators (SPIs)}\label{sec:dynamic_sc}
ML-based systems are often deployed in unpredictable environments which cannot be completely specified at design time. There is thus a need to continuously reassess the validity of the argument throughout the lifecycle. One way of incorporating runtime information into assurance arguments is through the use of Safety-Performance Indicators (SPIs), a concept first introduced in the context of autonomous vehicles by the UL 4600 standard \cite{UL4600_2023}. UL 4600 makes a distinction between traditional key performance indicators (KPIs) and SPIs, which are defined as performance metrics \textit{specifically related to safety}. SPIs are defined to be ``metrics supported by evidence that use threshold comparisons of condition claims in a safety case''. These metric values are obtained empirically by monitoring the system over its lifecycle. Each SPI is associated with a specific claim in the safety case, and if an SPI's threshold is violated (meaning the SPI is false), it indicates a violation of the claim. The standard further distinguishes between \textit{lagging} and \textit{leading} SPIs. Lagging SPIs are inherently reactive and might include metrics such as the number of accidents, instances of system failures, or the frequency of interventions by human drivers. On the other hand, leading SPIs focus on proactive measures such as the number of low-confidence classifications. Integrating both lagging and leading SPIs allows for a balanced approach to safety management.

SPIs may impact confidence in the associated claim(s) both positively and negatively. A lack of SPI violations should, intuitively, \textit{strengthen} confidence in a claim. The amount to which this impact can be made explicit in the argument relies on the underlying confidence assessment method. For qualitative approaches, an SPI can only have a binary effect (i.e., refuting or leaving the claim unchanged). Expressiveness is therefore limited and more complex effects that operational counter-evidence provided by the SPI might have on the claims (as, e.g., described in \cite{hawkinsIdentifyingRunTimeMonitoring2023}) cannot be made explicit. Instead, sustained performance can only be indirectly acknowledged through periodic qualitative reviews or expert assessments that affirm the system's ongoing safety. In the quantitative case, the impact can be modelled more explicitly, dependent upon the formalism being used. For example, if plain probabilities are used to quantify confidence in claims, then SPIs may either increase or decrease the probability (depending on whether they provide positive or negative evidence at runtime). In the case of formalisms such as Dempster-Shafer belief functions or Subjective Logic (SL) which allow for the explicit representation of uncertainty, the impact of SPIs can be modelled in even more subtle ways, as shown for the related concept of defeaters in  \cite{herd2025}.   

It is useful to distinguish an SPI’s \emph{update pattern} from its \emph{role}. With respect to the update pattern, SPIs can be (i) runtime, producing windowed evidence streams (e.g., perception error rates), (ii) event-driven, updating on specific triggers (e.g., major software releases), or (iii) periodic/static, re-evaluated infrequently (e.g., judging data representativeness). With respect to role (as already mentioned above), SPIs can be leading -- producing early, actionable indicators such as perception monitor violations -- or lagging -- producing outcome rates grounded in accumulated statistics such as incident rates. This paper focuses on leading, windowed runtime SPIs; the update rule presented in this paper (Eq.~\eqref{eq:update_single}) applies to those windows and supports immediate penalties on violations. Lagging SPIs are updated less frequently when fresh outcome data are available. Between updates, the associated assessments remain unchanged or may be conservatively down-weighted to prevent stale evidence from dominating.

\subsection{Subjective Logic}\label{sec:subjective_logic}

Subjective Logic (SL) \cite{josang2016subjective} is a framework for reasoning with uncertain beliefs that combines ideas from probabilistic logic and evidence theory. The atomic building blocks of SL are \emph{subjective opinions}, and SL offers a range of combination operators that allow for algebraic reasoning. Subjective opinions express beliefs about the truth of propositions under degrees of uncertainty. Throughout this paper, we focus on \emph{binomial} opinions, i.e.\ opinions about a binary domain $X=\{x,\bar{x}\}$, since safety assurance claims are typically binary statements (the claim holds or it does not). Likewise, we treat SPIs as binary predicates derived from thresholded measurements: an SPI is satisfied or violated based on the observed metric relative to its threshold, while the amount and consistency of measurement data are reflected in the associated belief, disbelief, and uncertainty (and, via the opinion–Beta mapping described below, in the distribution’s width).

\begin{definition}[Binomial opinion]\label{def:binop}
Let $\mathbb{X} = \{x,\bar{x}\}$ be a binary domain. A binomial opinion about the truth of $x$ is a tuple $\omega_x = (b_x,d_x,u_x,a_x)$ where 
\begin{itemize}[noitemsep,topsep=5pt]
    \item $b_x$ (belief): the belief mass in support of $x$ being \textbf{\textit{true}}
    \item $d_x$ (disbelief): the belief mass in support of $x$ being \textbf{\textit{false}}
    \item $u_x$ (uncertainty): the uncommitted belief mass
    \item $a_x$ (base rate): the \textit{a priori} probability in the absence of committed belief mass (often set to 0.5 for binary domains) 
\end{itemize}
with $b_x,d_x,u_x,a_x \in [0,1]$ and $b_x + d_x + u_x = 1$. 
\end{definition}

\subsubsection{Constructing opinions:} Given positive evidence $r$ (number of positive observations)  for a claim, negative evidence $s$ (number of negative observations) and a \textit{non-informative prior weight}\footnote{$W$ ensures that when evidence begins to accumulate (i.e.\ $r$ gets larger), uncertainty $u_x$ decreases accordingly. $W$ is typically set to the same value as the cardinality of the domain (2 in our binary case), thus artificially adding one success $r$ and one failure $s$. Higher values of $W$ require more evidence for uncertainty to decrease.} $W$, a binomial opinion can be computed as follows: 
\vspace{-0.1cm}
\begin{align}
    b_x &= r/(r+s+W)\label{eq:belief}\\
    d_x &= s/(r+s+W)\label{eq:disbelief}\\
    u_x &= W/(r+s+W)\label{eq:uncertainty}
\end{align}
with base rate $a\in[0,1]$. $W$ controls how fast uncertainty $u$ decreases as evidence accumulates (commonly $W=2$ in binary domains). Binomial opinions correspond to Beta distributions: with $r$, $s$, $a$, and $W$, the corresponding Beta parameters are
$\alpha = r + aW$ and $\beta = s + (1-a)W$, and $E(x)=\alpha/(\alpha+\beta)=(r+aW)/(r+s+W)$. For binary domains with $a=0.5$ and $W=2$, the opinion--Beta mapping corresponds to Laplace's rule of succession with a Beta$(1,1)$ prior, yielding $E(x)=(r+1)/(r+s+2)$; we adopt this non-informative prior unless domain knowledge justifies a different $(a,W)$.

\subsubsection{Confidence:} In SL, a binomial opinion $\omega$ expresses how much is \emph{believed}, \emph{disbelieved}, or remains \emph{uncertain} about a claim. One of the strengths of SL is to distinguish between two complementary notions of confidence:

\begin{itemize}
    \item \textbf{First-order confidence} describes how much of the opinion is \emph{committed} (either to belief or disbelief) rather than left uncertain. Practically, it increases as uncertainty $u$ shrinks -- e.g., when more consistent evidence is observed or independent sources are fused.
    \item \textbf{Second-order confidence} describes how \emph{precise} that commitment is. As described above, every opinion corresponds to a Beta distribution; a narrow, peaked Beta reflects high second-order confidence, while a wide, flat Beta reflects low second-order confidence. Second-order confidence grows with more independent observations and is influenced by the chosen prior weight $W$. 
\end{itemize}

\subsubsection{Negating opinions:}

Negation in SL for binomial opinions swaps belief and disbelief while complementing the base rate. For an opinion $\omega_x = (b_x, d_x, u_x, a_x)$ about proposition $x$, the negated opinion is $\lnot \omega_x \equiv \omega_{\bar{x}} = (d_x, b_x, u_x, 1-a_x)$. Intuitively, belief that $x$ holds becomes belief that $x$ does not hold, while uncertainty remains unchanged. We use this formal negation in Eq.~\eqref{eq:update_single} to ensure that belief in an SPI violation (i.e., belief that `SPI is false’) correctly maps to disbelief in its associated claim.

\subsubsection{Combining opinions:}\label{sec:combining} SL provides a wide range of combination operators \cite{josang2016subjective} that offer an elegant and intuitive way to combine opinions instead of the underlying Beta distributions, a direct manipulation of which would be mathematically challenging. In this work, we use two operators -- \textit{cumulative fusion} and \textit{refuting challenger} -- and combine them as described in Section\ \ref{sec:methodology}. 

\paragraph{Cumulative Belief Fusion (CBF)} CBF applies when independent evidence about the \textit{same} claim accumulates, so uncertainty decreases as more independent observations are added (from more or the same sources). Given two independent binomial opinions $\omega^A_x$ and $\omega^B_x$ held by sources $A$ and $B$ on the same binary domain $\mathbb{X}=\{x,\bar{x}\}$, their cumulative fusion is $\omega^A_x \oplus \omega^B_x$. 

\paragraph{Refuting challenger (RC)} The RC is one of two \textit{challenge} operators (\textit{SC=skeptical challenger} and \textit{RC=refuting challenger}) introduced in \cite{herd2025} that represent the effect of challenging the validity of a target opinion $\omega^B_x$ on the binary domain $\mathbb{X}=\{x,\bar{x}\}$ using a challenger opinion $\omega^A_B$, denoted as $ \omega^A_B \boxtimes \omega^B_x$ ($\omega^A_B$ challenges $\omega^B_x$). Intuitively, increasing belief in $\omega_s$ reallocates committed mass in $\omega_t$ away from belief. In this paper, we use RC to achieve an \textit{invalidating} effect, i.e., to move this mass to disbelief while leaving uncertainty $u$ unchanged. This reflects the idea that an SPI violation provides \textit{specific counter‑evidence} rather than increased uncertainty.

\subsubsection{SL versus classical probabilistic updates:}

It is important to note that SL does not replace classical probability theory. While second-order uncertainty can also be represented in probability theory via distributions over parameters, SL provides an explicit, compositional calculus over belief, disbelief, and uncertainty with a direct mapping to Beta distributions, including operators that preserve this structure across the argument. Thus, even when claims and their updates are driven by numeric measurements (e.g., SPI counts or ML performance metrics), SL offers practical benefits for assurance confidence quantification: claim-level visibility of first- and second-order uncertainty, standardized composition and negation, and auditable update semantics. In particular, while the special case $a=0.5$, $W=2$ in binomial domains recovers Laplace-smoothed probabilities at the level of expectations, SL retains explicit uncertainty mass $u$ and algebraic operators for composition and challenge. Furthermore, SL supports explicitly non-conjugate, operator-level updates: Eq.~\eqref{eq:update_single} described below combines cumulative fusion with a refuting challenger on the negated SPI opinion to impose an immediate penalty at the moment of violation. This operator-level reallocation of mass departs from conjugate Beta-Binomial updates, which add counts and typically increase confidence. Meta-uncertainty about priors or model choice can, in principle, also be represented in SL (e.g., via trust discounting), but we consider this out of scope here.
\section{SL-based dynamic confidence updates}\label{sec:methodology}

\begin{figure*}[tbp!]
        \centering
        \resizebox{0.8\textwidth}{!}{\begin{tikzpicture}[
        module/.style={draw, thick, rounded corners, minimum width=15ex, font=\sffamily},
        argument/.style={module, solid, thin, inner sep=10pt},
        claim/.style={module, fill=cyan!20, align=center},
        evidence/.style={module, fill=blue!5, ellipse},
        spi/.style={module, fill=yellow!20},
        arrow/.style={-stealth', thick, rounded corners}, 
        dotted_arrow/.style={dashed, ->, thick}
    ]
        \node[claim] (claim) {
        \text{Claim $c$}};
        
        \node[claim, fill=white, right =of claim] (opp_claim) {\begin{tikzpicture}[scale=0.4, transform shape]
            \begin{axis}[
                width=4cm, height=3cm, 
                axis lines=middle, 
                xtick=\empty, ytick=\empty, 
                enlargelimits=false,
                xlabel style={at={(ticklabel* cs:1.05)}, anchor=north},
                ylabel style={at={(ticklabel* cs:1.05)}, anchor=south}
            ]
                \addplot[teal, thick, domain=0:1, samples=1000] {2*x*(1-x)^5};
            \end{axis}
        \end{tikzpicture}};
        
        \draw[dotted_arrow] (claim) -- (opp_claim);
        
        \node[evidence, text width = 2cm, below=of claim] (design_evidence) {Design time Evidence};
        \draw[arrow] (claim) -- (design_evidence);
        
        \node[argument, fit=(claim) (design_evidence) (opp_claim), label={[anchor=north, yshift=5mm]above:Design-time}] (design_box) {};
        
        \node[spi, right=10cm of opp_claim] (spi) {$SPI_c$};
        \node[claim, fill=white, left=1cm of spi] (opp_spi) {\begin{tikzpicture}[scale=0.4, transform shape]
            \begin{axis}[
                width=4cm, height=3cm, 
                axis lines=middle, 
                xtick=\empty, ytick=\empty, 
                enlargelimits=false,
                xlabel style={at={(ticklabel* cs:1.05)}, anchor=north},
                ylabel style={at={(ticklabel* cs:1.05)}, anchor=south}
            ]
                \addplot[red, thick, domain=0:1, samples=1000] {2*x*(1-x)^2};
            \end{axis}
        \end{tikzpicture}};

        \node[spi, below left=of opp_spi, font=\large](update){$\omega'_c = {\neg \omega^{SPI}_{c}} \boxtimes (\omega_{c} \oplus {\omega^{SPI}_{c}})$};

        \draw[dotted_arrow] (opp_spi.south) |- (update.east) node[fill=green!10!white,midway, above, font=\large] {${\omega^{SPI}_{c}}$};
        \draw[arrow] (update.north) |- (opp_claim.east) node[fill=white, midway, above, font=\large] {$\omega'_c$};
        \draw[dotted_arrow] (opp_claim.south) |- (update.west) node[fill=white,midway, above, font=\large] {${\omega_{c}}$};

        \draw[dotted_arrow] (spi) -- (opp_spi);
    
        \node[evidence, text width = 1.5cm, below=of spi] (runtime_evidence) {Runtime Evidence};
        \node[argument, fill=green!10!white, solid, text width = 2.2cm, right=1cm of runtime_evidence,label={[anchor=north, yshift=5mm]above:Data source}] (runtime) {Simulation,\\ Deployment, ...};
        \draw[arrow] (spi) -- (runtime_evidence);
        \draw[arrow] (runtime) -- (runtime_evidence);
        
        \begin{scope}[on background layer]
        \node[argument, fill=green!10!white, solid, fit=(spi) (opp_spi) (runtime_evidence), label={[anchor=north, yshift=5mm]above:SPI Monitoring}] (runtime_box) {};
        \end{scope}
        
        \node[argument, inner sep=15pt, fit=(runtime_box)(runtime), label={[anchor=north, yshift=5mm]above:Runtime}] (runtime_time_box) {};

    \end{tikzpicture}}
    \caption{Relationship between claims, SPIs, their associated opinions, and the confidence update process}
\label{fig:overview}
\end{figure*}
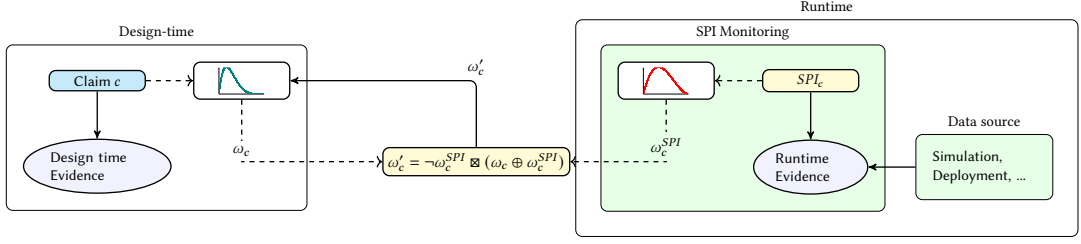

We propose a methodology for integrating runtime information into SL-based quantitative assurance arguments. An overview of this methodology is shown in Fig.~\ref{fig:overview}. The starting point is a structured design-time assurance argument with \textit{claims} and \textit{sub-claims} supported by \textit{evidence}. We associate each claim $c$ in the argument with a binomial opinion $\omega_c$ that represents belief, disbelief, and uncertainty in $c$ in a quantitative way. $\omega_c$ is either derived from concrete objective evidence or formed based on a subjective judgment of the claim's validity. Runtime information is then integrated into the design-time argument as follows: 

\begin{enumerate}
    \item Where relevant, claims $c$ in the argument are associated with Safety Performance Indicators $SPI_c$. SPI monitors are then derived, collecting runtime information to validate $c$.  
    \item Each $SPI_c$ is associated with a binomial opinion $\omega^{SPI}_c$ about the windowed predicate ``SPI holds''. Within a window of $k$ observations, $r$ successes (frames satisfying the SPI threshold) and $s$ failures (violations) are counted and composed into $\omega^{SPI}_c$ via Eqs.~\eqref{eq:belief}--\eqref{eq:uncertainty}. Thus $\omega^{SPI}_c$ does not describe a single observation being accepted or rejected; it summarizes the evidence about the predicate over the window. Uncommitted belief $u$ captures ignorance due to limited evidence (small $r{+}s$ ), and it shrinks as more observations accumulate. 
    \item $\omega_c$ is updated with $\omega^{SPI}_c$ to reflect the updated confidence. To this end, \textit{cumulative fusion} is combined with a \textit{refuting challenger} (Sec.~ \ref{sec:combining}). The combination ensures that \textit{positive evidence} for the SPI (lack of SPI violations) increases confidence in the original claim, whereas \textit{negative evidence} (SPI violations) causes confidence in the original claim to drop.  
\end{enumerate} 
We assume that (i) SPI windows are non-overlapping and independent, and (ii) claim and SPI opinions use a consistent base rate $a$ and prior weight $W$. We describe our proposed methodology in more detail below, using the  running example of an ML-based construction cone detection model. 

\subsection{Running example: ML-based construction cone detection}\label{sec:running_example}

We illustrate our approach using an ML-based \emph{Construction Zone Assist} (CZA) function which is responsible for the reliable detection of highway construction sites. We focus here on the perception module. Construction areas pose significant challenges because of their unpredictable setup, which can involve irregular road configurations, temporary signs, as well as both static and dynamic obstacles like heavy machinery and workers. 

In a simulated environment, we execute a cone detection component within an overall construction zone detection sub-system using APIKS (Autonomous Platform at IKS\footnote{\scriptsize{\url{https://www.iks.fraunhofer.de/en/services/apiks-software-platform-for-autonomous-vehicle-functions.html}}}), a ROS2-based \cite{ros2} autonomous vehicle software platform. An integrated perception module leverages a YOLOv8 \cite{yolov8_ultralytics} object detector; the design-time confusion matrix results of this cone detection component provide an initial assessment of the model’s performance claim. APIKS is used to simulate various runtime scenarios in the CARLA simulator \cite{carla}, including construction sites in different environmental conditions. Ground truth and predicted bounding boxes of traffic cones are used to compute runtime evidence.

\subsection{Design-time argument}

\begin{figure*}[tbp!]
        \centering
        \resizebox{0.8\textwidth}{!}{\input{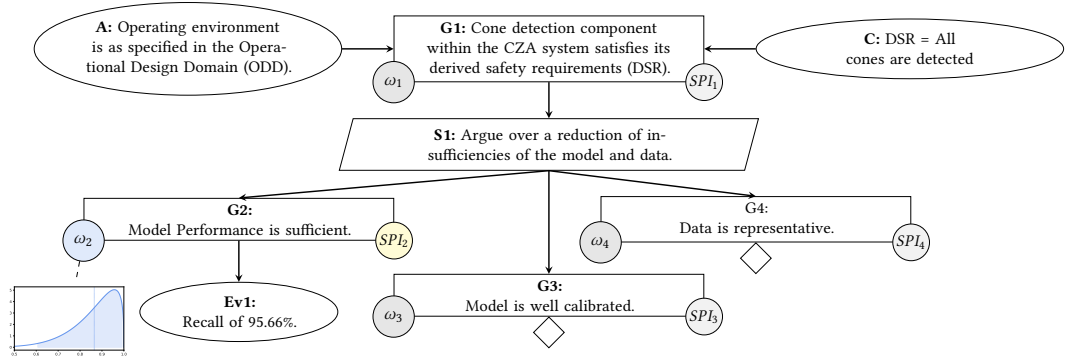}}
        \label{fig:rc}
    \caption{Example GSN assurance argument for the ML cone detection component}
    \label{fig:gsn_example}
\end{figure*}

We use a simple design-time argument as an exemplary starting point for analysis, shown in GSN\footnote{\scriptsize{Goal Structuring Notation; \url{https://scsc.uk/GSN}}} in Fig. \ref{fig:gsn_example}. The argument aims to justify the safety of the ML-based cone detection component (represented by the safety requirement in context node $C$) by arguing over a reduction of insufficiencies on the model and data level. Each claim $c$ is associated with a binomial opinion (see Def.~\ref{def:binop}) that expresses confidence in $c$. 

We focus on claim \textbf{G2} which argues about model performance by using a test-time recall value of 95.66\% as evidence\footnote{Other claims can be dealt with accordingly.}, computed based on 106 false negatives (FNs) and 2334 true positives (TPs). Using Eqs.\ \eqref{eq:belief}--\eqref{eq:uncertainty} with $r=TP$ and $s=FN$, we form an opinion $\omega_2 = (b: 0.9558, d: 0.0434, u: 0.0008)$ which expresses high certainty due to the large sample size. To account for additional uncertainty in this measurement, e.g., due to insufficient knowledge about the quality of the dataset, we inject additional uncertainty mass into $\omega_2$. The corresponding Beta distribution is shown in Fig. \ref{fig:gsn_example} below $\omega_2$.

\subsection{Integration of SPIs}

The static design-time argument is extended with SPIs to incorporate runtime information. Each SPI is linked with one or several argument claims and is supported by evidence obtained at runtime through an \textit{SPI monitor}. The SPI monitor logic for \textbf{G2} is shown in Fig. \ref{fig:spi_monitor}. It computes, for each frame in the simulation, the ratio of false negative cone detections to the actually present cones\footnote{Since monitoring happens in a simulation environment, ground truth is available.} within distance $d$. If the ratio exceeds threshold $\theta$, it counts as an SPI violation.

\begin{figure}[tbp!]
\centering
\begin{tcolorbox}[colframe=black, colback=white]
\textbf{Parameters:} SPI window size $k$, max.\ detection distance $d$, threshold $\theta$

\textbf{Function} $\text{SPI}(frames)$:
\begin{enumerate}
    \item $r:= 0$, $s:= 0$, $W := 2$\label{lst:step1}
    \item \textbf{For each} frame in $k$ frames:\label{lst:step2}
    \begin{enumerate}
        \item $\text{FN} = \text{CountFalseNegatives(frame, } d\text{)}$
        \item $\text{GT} = \text{CountGroundTruthObjects(frame, } d\text{)}$
        \item $\text{ratio} = \frac{\text{FN}}{\text{GT}}$ \textbf{if} $GT >0$ \textbf{else} $0$
        \item \textbf{If} $\text{ratio} \geq \theta$ \textbf{Then} $s \mathrel{+}= 1$ \textbf{Else} $r \mathrel{+}= 1$
    \end{enumerate}
    \item Compute resulting SPI opinion:\label{lst:step3}
    \begin{itemize}
        \item $b=r/(r+s+W)$
        \item $d=s/(r+s+W)$
        \item $u=W/(r+s+W)$
    \end{itemize}
    \item \textbf{Return} $\omega^{SPI}_c = (b,d,u,a=0.5)$\label{lst:step4}
\end{enumerate}
\end{tcolorbox}
\caption{Logic of the SPI monitor in pseudocode form}
\label{fig:spi_monitor}
\end{figure}

Similarly to design-time claims, each SPI is associated with a binomial opinion that is derived from the runtime evidence. The actual derivation is dependent upon the nature of the SPI and its threshold. The SPI monitor for \textbf{G2} in the example uses a window of size $k=10$ frames and threshold $\theta = 0.5$: the opinion is constructed by counting the SPI violations within the SPI window as failures $s$ and the number of frames without violations as successes $r$, and applying Eqs.~\eqref{eq:belief}--\eqref{eq:uncertainty} (see steps \ref{lst:step3} and \ref{lst:step4} in Fig.~\ref{fig:spi_monitor}). This produces a new SPI opinion $\omega^{SPI}_2$ for every non-overlapping window of $k$ frames which can then be used to update the original claim opinion $\omega_2$ as described next.

\subsection{Confidence update}

The central step is the update of claim opinion $\omega_c$ with SPI opinion $\omega^{SPI}_c$. Because SPI violations are safety‑critical signals that must trigger prompt corrective actions, while the absence of violations constitutes only weak, accumulating evidence, we require the update to satisfy two properties: 

\begin{enumerate}
    \item \textbf{Confirmation:} When no SPI violations occur, then confidence that claim $\omega_c$ holds should increase gradually, reflecting the idea that a lack of violations can be seen as an accumulation of positive evidence.  
    \item \textbf{Violation penalty:} Any SPI violation should have an immediate negative effect on confidence in $\omega_c$. This reflects the idea that any SPI violation indicates a potential invalidation of the underlying claim.
\end{enumerate}

To achieve the two properties, we combine the cumulative fusion (CBF) and the refuting challenger (RC) operator to produce an updated claim opinion $\omega'_c$: 

\begin{align}\label{eq:update_single}
    \omega_c' &= \neg \omega^{SPI}_c \boxtimes (\omega_c \oplus \omega^{SPI}_c)
\end{align}
where $\oplus$ is CBF and $\boxtimes$ is RC. Here $\omega^{\mathrm{SPI}}_c$ is built from the window’s total evidence (successes $r$ and failures $s$) using Eqs.~\eqref{eq:belief}--\eqref{eq:uncertainty}. The inner CBF accumulates this evidence and reduces uncertainty as more runtime data is observed. The outer RC then applies an \emph{immediate refutation} proportional to the (negated) SPI belief, which represents the direct effect of observed violations on the claim in the same update step. Note that negation is applied so that belief in the challenger maps to disbelief in the target claim. 

The design satisfies the two properties above: (i) when SPI windows show no violations, belief in the claim increases and uncertainty decreases via CBF; (ii) violations cannot increase claim belief, since RC reallocates mass from belief to disbelief proportional to the (negated) SPI belief. The update deliberately departs from the standard conjugate Beta-Binomial Bayesian update: the combination of cumulative fusion with a refuting challenger reallocates belief mass in a way that, whenever violations occur, cannot be obtained by simply adding Bernoulli counts; it imposes a targeted refutation to reflect the immediate impact of SPI violations.

\subsection{Experimental results} 

\newcommand{\svgh}{2.8cm}

\begin{figure}[tbp!]
    \centering
    \setlength{\tabcolsep}{1pt}     
    \renewcommand{\arraystretch}{0} 

    \begin{tabular}{cc}
        \subfloat[]{%
            \adjustbox{padding=6pt}{\includegraphics[width=0.48\linewidth]{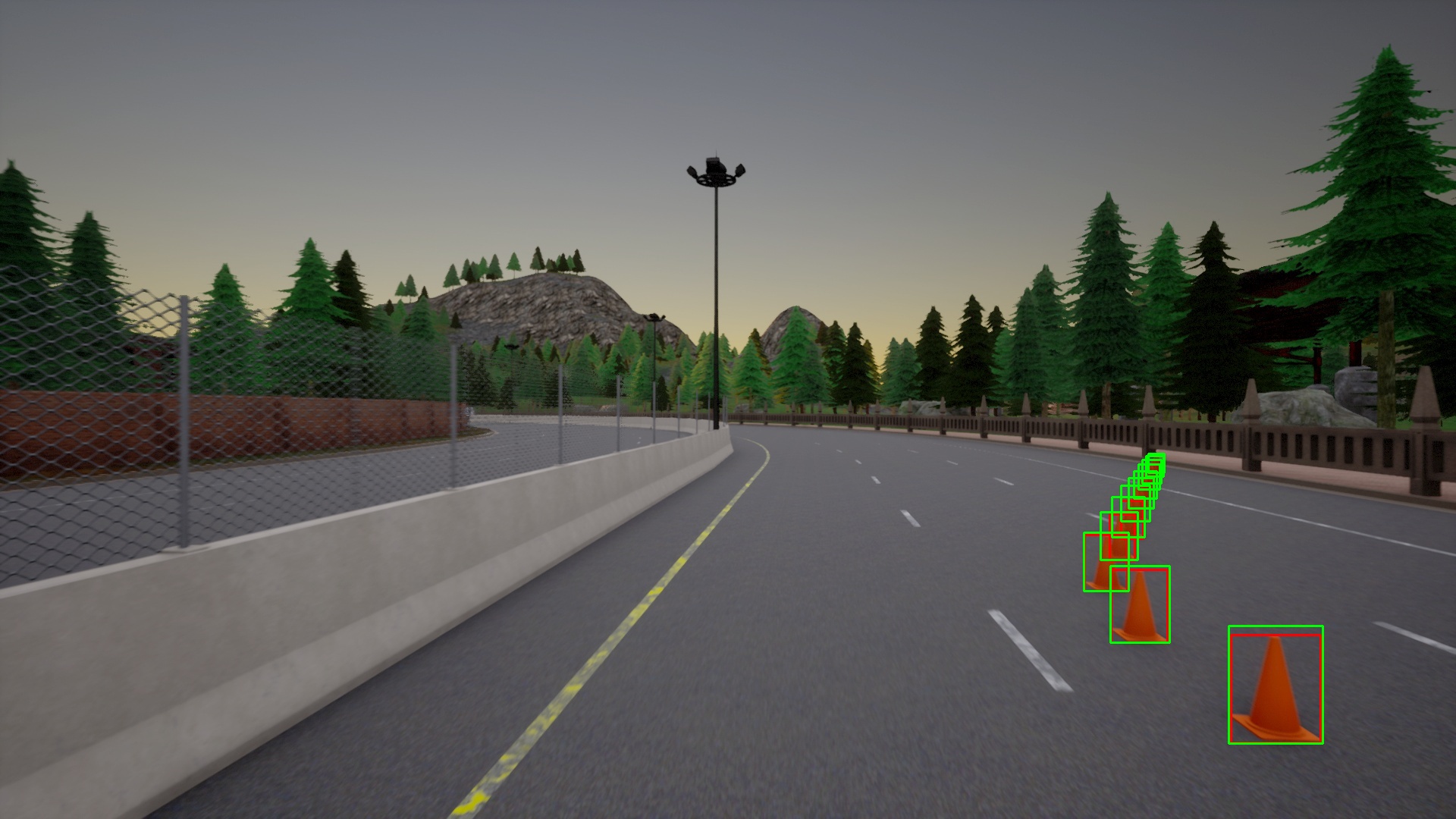}}%
        } &
        \subfloat[]{%
            \includegraphics[width=0.48\linewidth, height=\svgh, keepaspectratio=false]{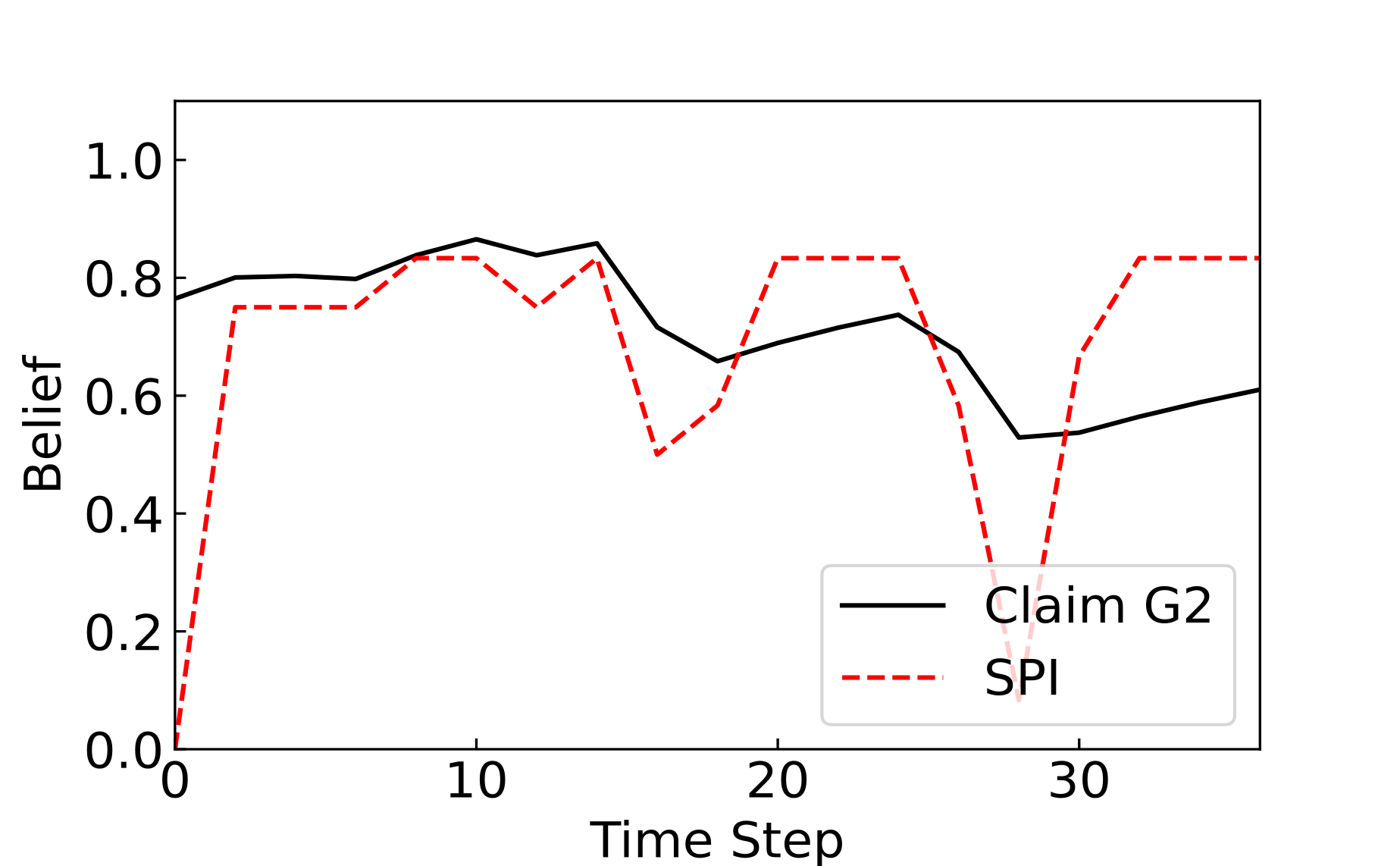}%
        } \\

        \subfloat[]{%
            \includegraphics[width=0.48\linewidth, height=\svgh, keepaspectratio=false]{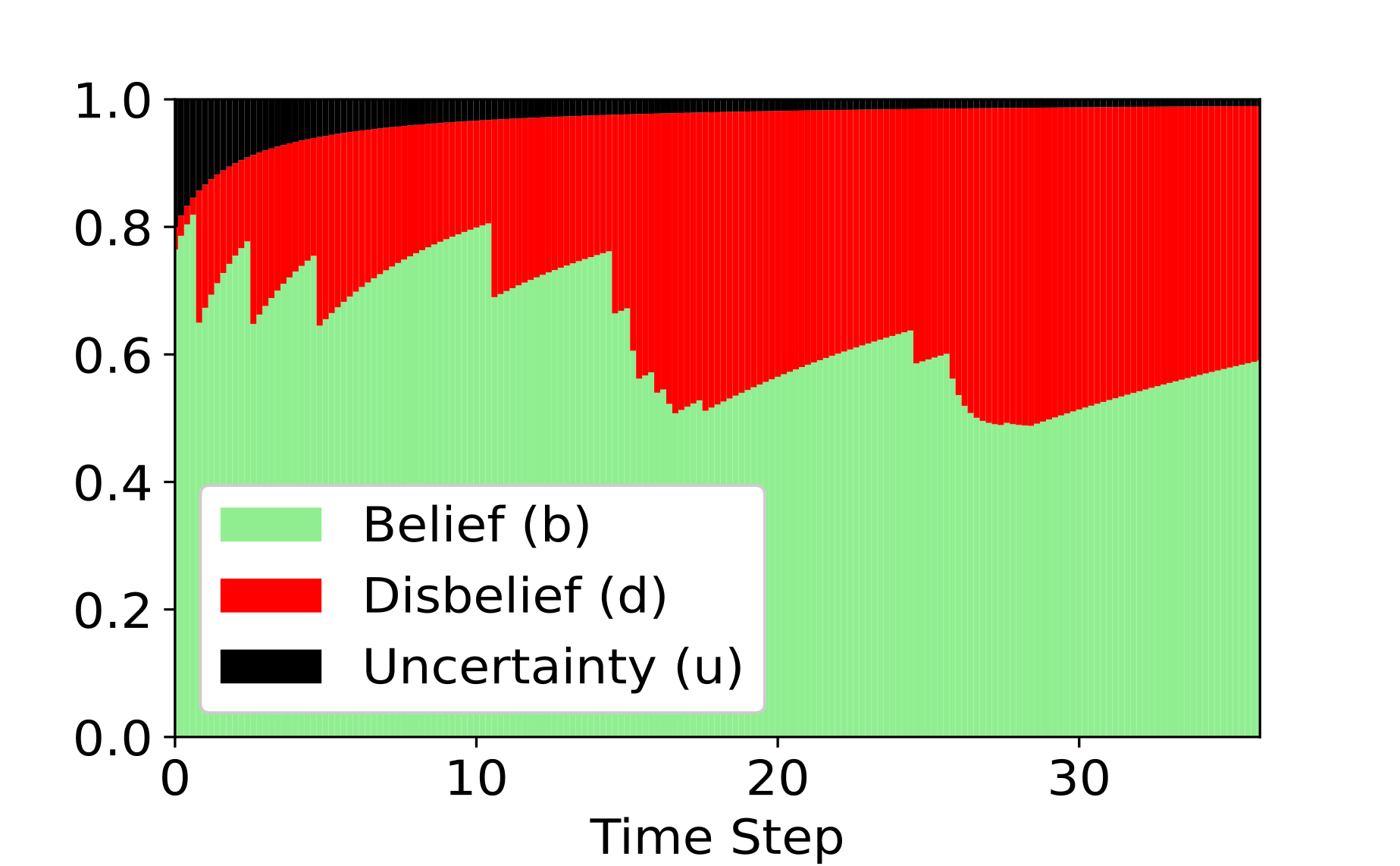}%
        } &
        \subfloat[]{%
            \includegraphics[width=0.48\linewidth, height=\svgh, keepaspectratio=false]{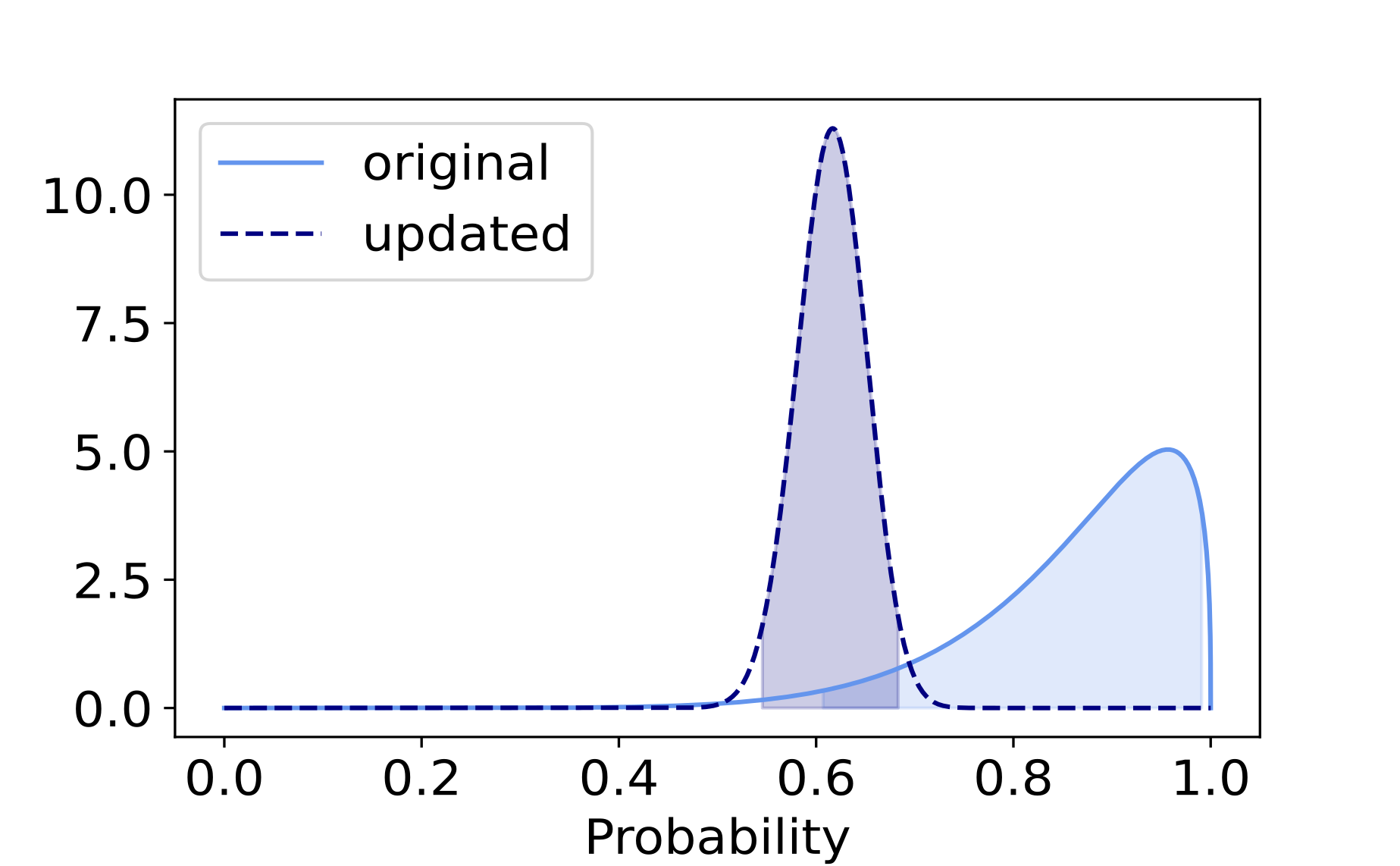}%
        } \\
    \end{tabular}
    \caption{Simulation scenario with an SPI violation showing (a) a simulation snapshot; (b) evolution of confidence in the SPI and claim opinion; (c) evolution of belief, disbelief, and uncertainty of the claim opinion; (d) belief in the claim at the beginning and the end of the scenario.}
    \label{fig:ScenarioA}
\end{figure}

The effect of the SPI monitor is demonstrated on a sample scenario from the APIKS simulation in Fig.~\ref{fig:ScenarioA}. In the figure, we see a snapshot of a scene from the simulation where several cones are mispredicted, likely attributable to occlusion due to cone placement. On the top right, the belief of both the SPI $\omega^{SPI}_c$ and claim opinion $\omega_c$ are shown over simulation steps. It is apparent that a drop in SPI belief leads to a drop in claim belief, followed by a slow recovery as confidence in the SPI is re-gained. By taking a deeper look into the components of claim opinion $\omega_c$, the bottom left graph shows that uncertainty is significantly reduced over time as runtime information is collected and fused with the original opinion. We also see an increase in disbelief when SPI violations are observed at runtime. The bottom right graph shows the Beta distribution for the original opinion as well as the opinion at the end of the simulation scenario. We can see that the resulting opinion is more certain, given by the narrower distribution, and has lower belief due to the presence of SPI violations.

\begin{figure*}[tb]
    \centering
    \setlength{\tabcolsep}{1pt}  
    \begin{tabular}{ccc}
        \subfloat{\includegraphics[width=0.25\textwidth]{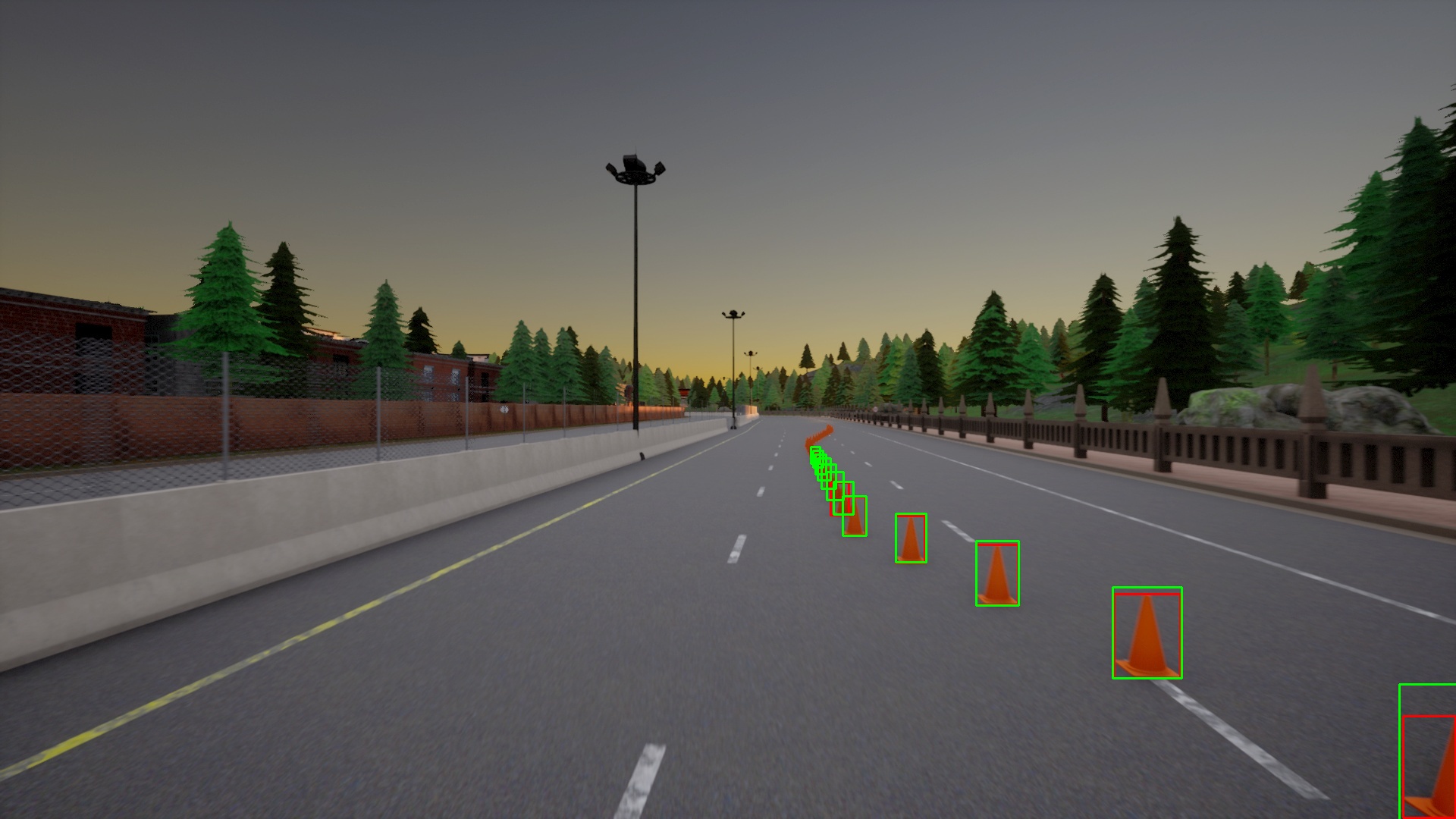}} &
        \subfloat{\includegraphics[width=0.25\textwidth]{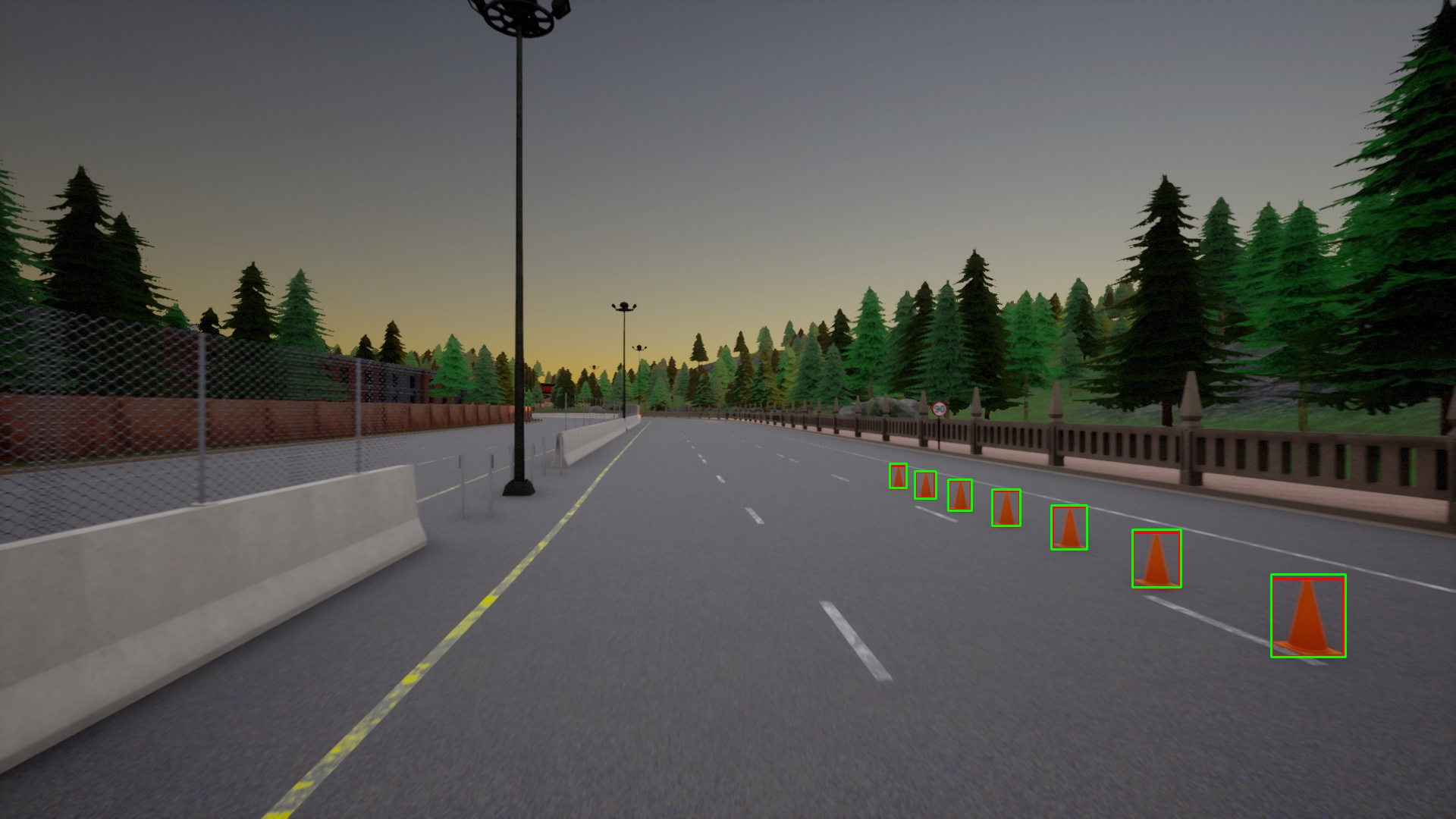}} &
        \subfloat{\includegraphics[width=0.25\textwidth]{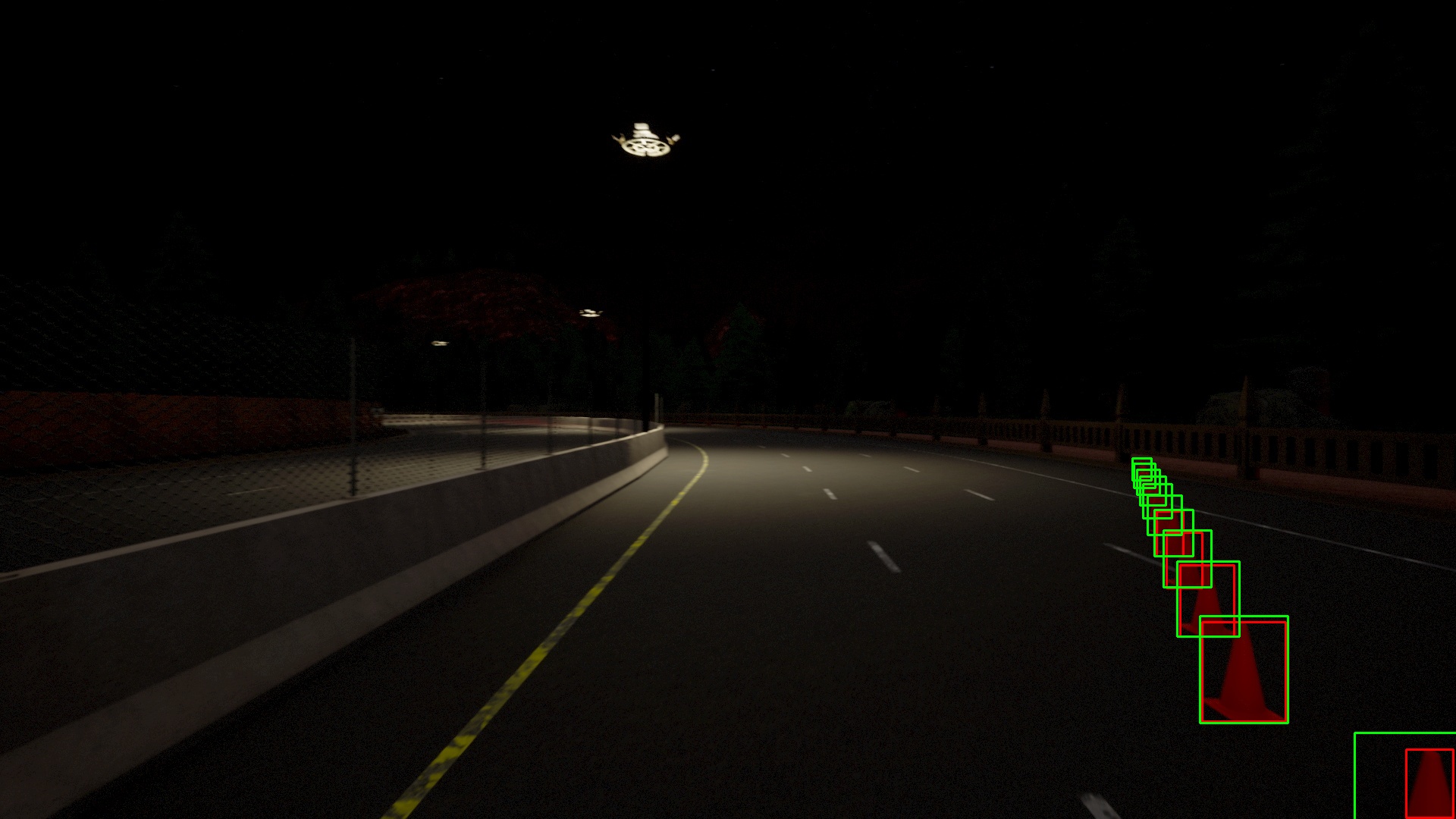}} \\
        \subfloat{\includegraphics[width=0.28\textwidth]{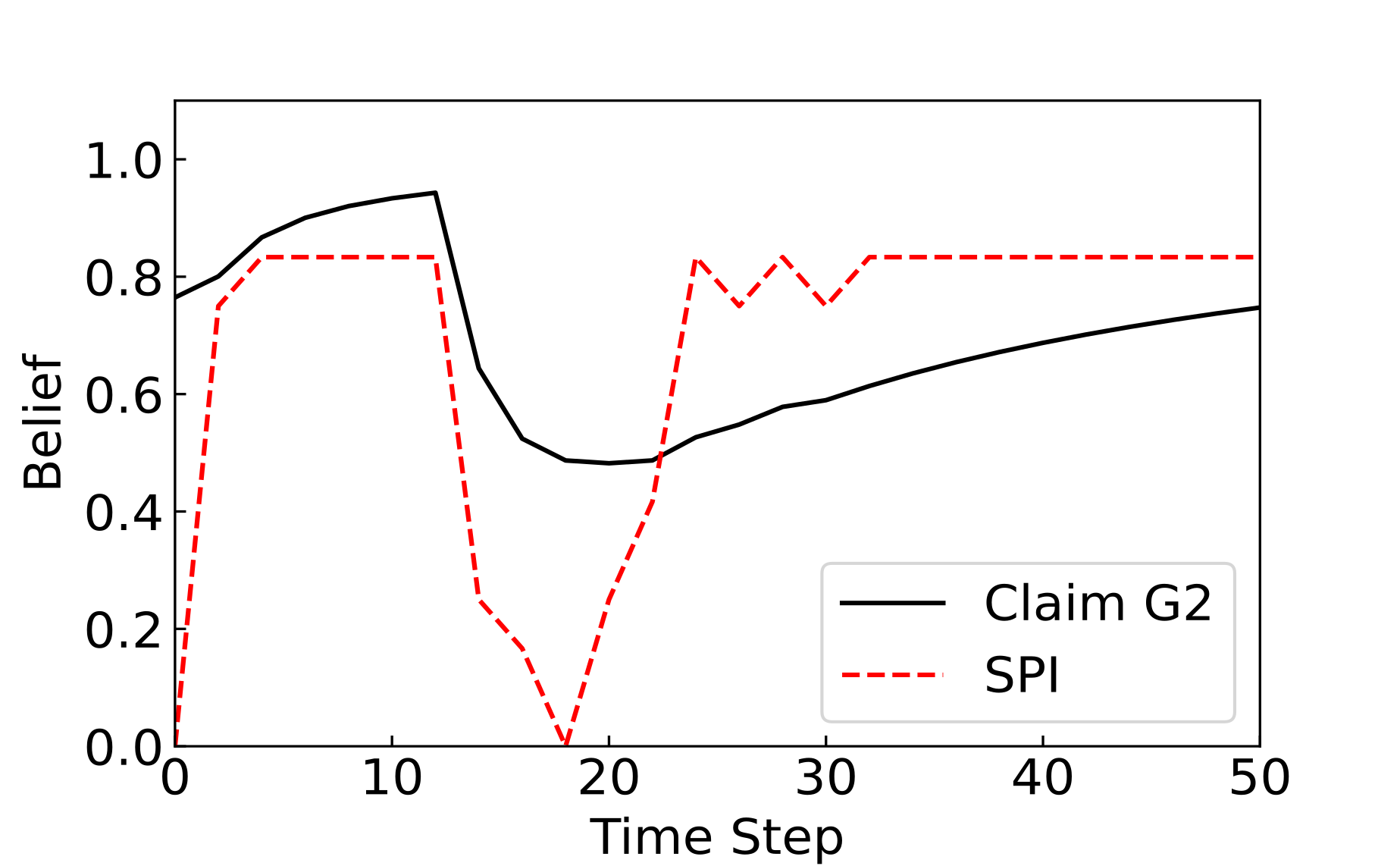}} &
        \subfloat{\includegraphics[width=0.28\textwidth]{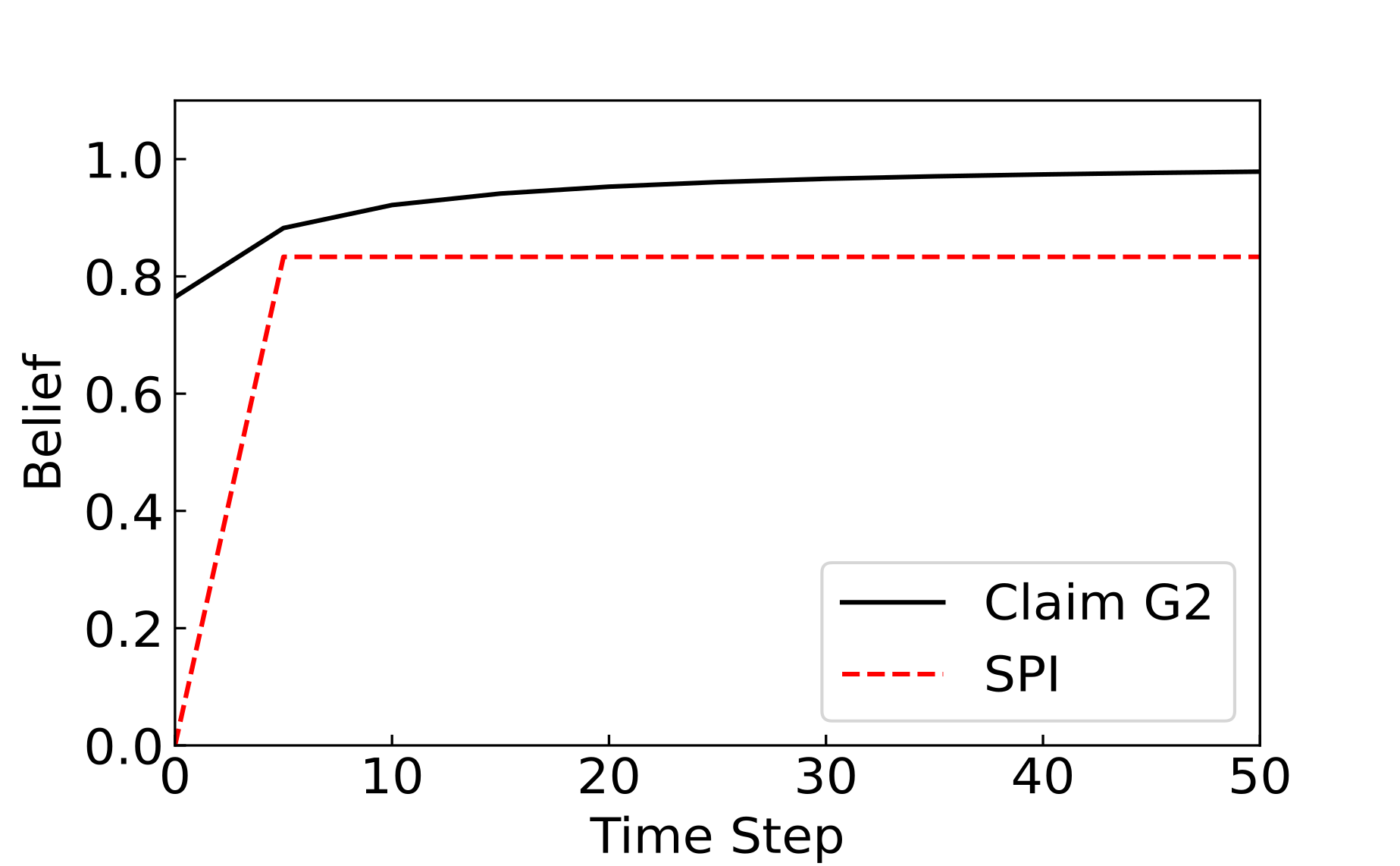}} &
        \subfloat{\includegraphics[width=0.28\textwidth]{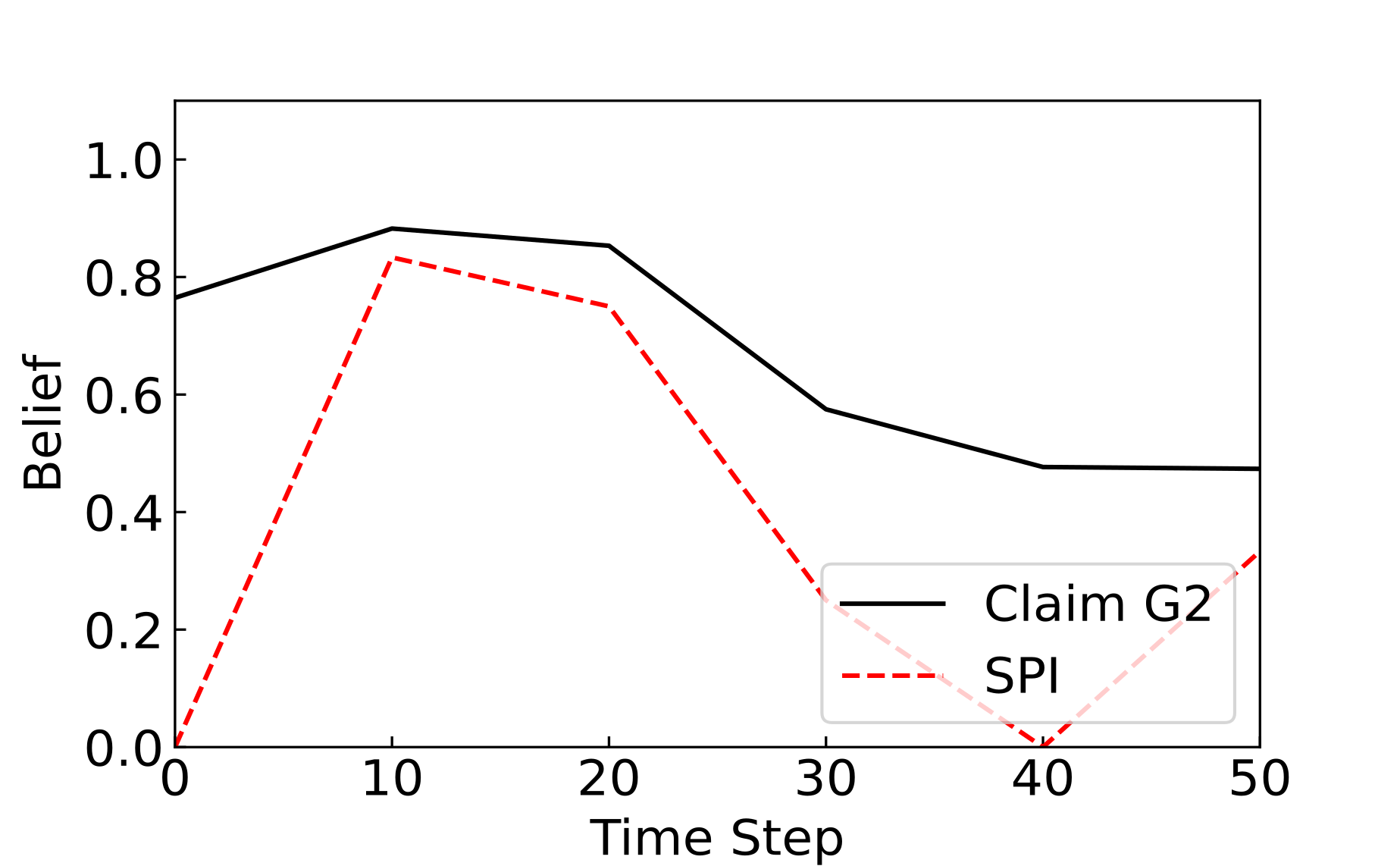}} \\
        \subfloat{\includegraphics[width=0.28\textwidth]{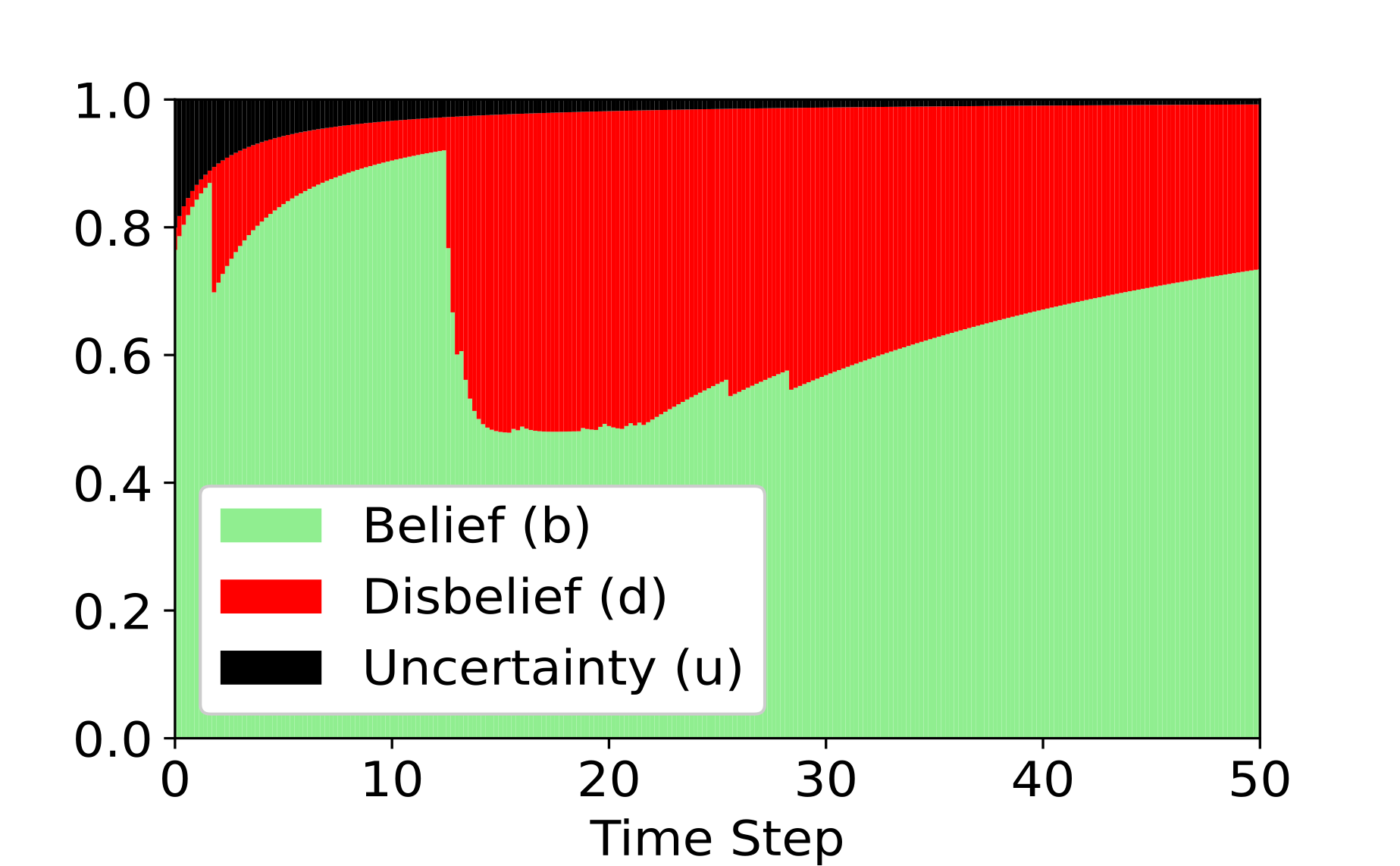}} &
        \subfloat{\includegraphics[width=0.28\textwidth]{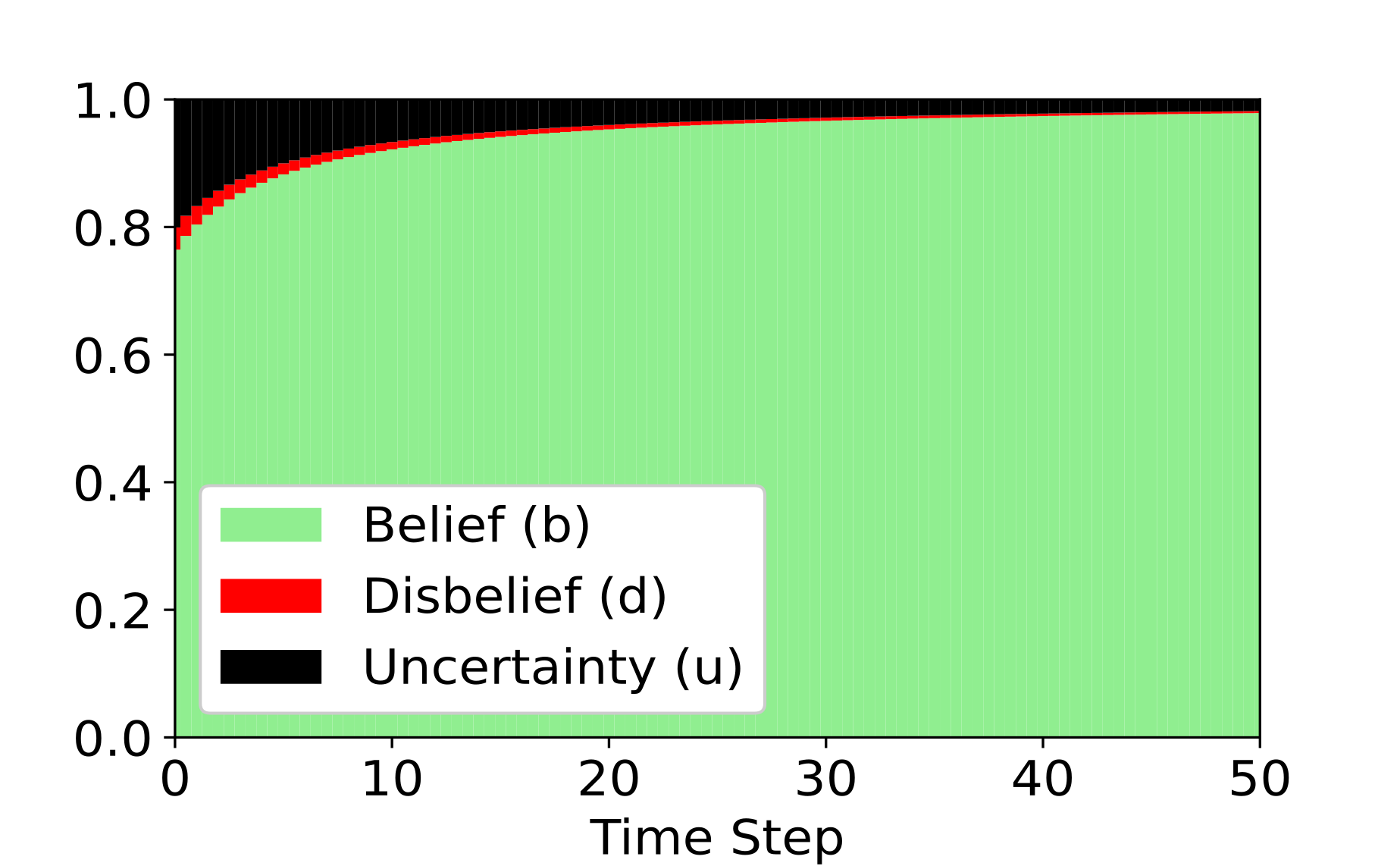}} &
        \subfloat{\includegraphics[width=0.28\textwidth]{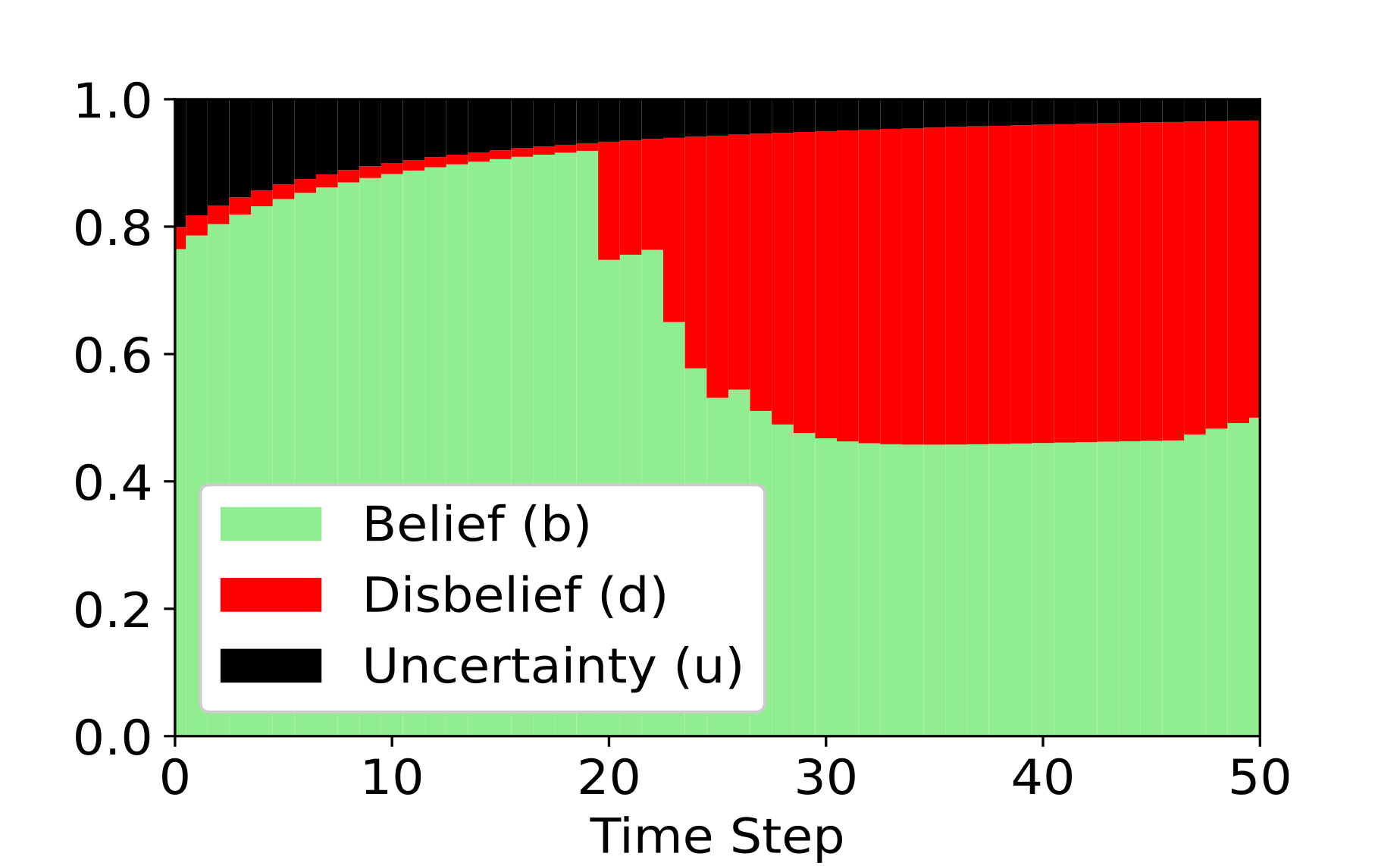}} \\
        \setcounter{subfigure}{0}
        \subfloat[]{\includegraphics[width=0.28\textwidth]{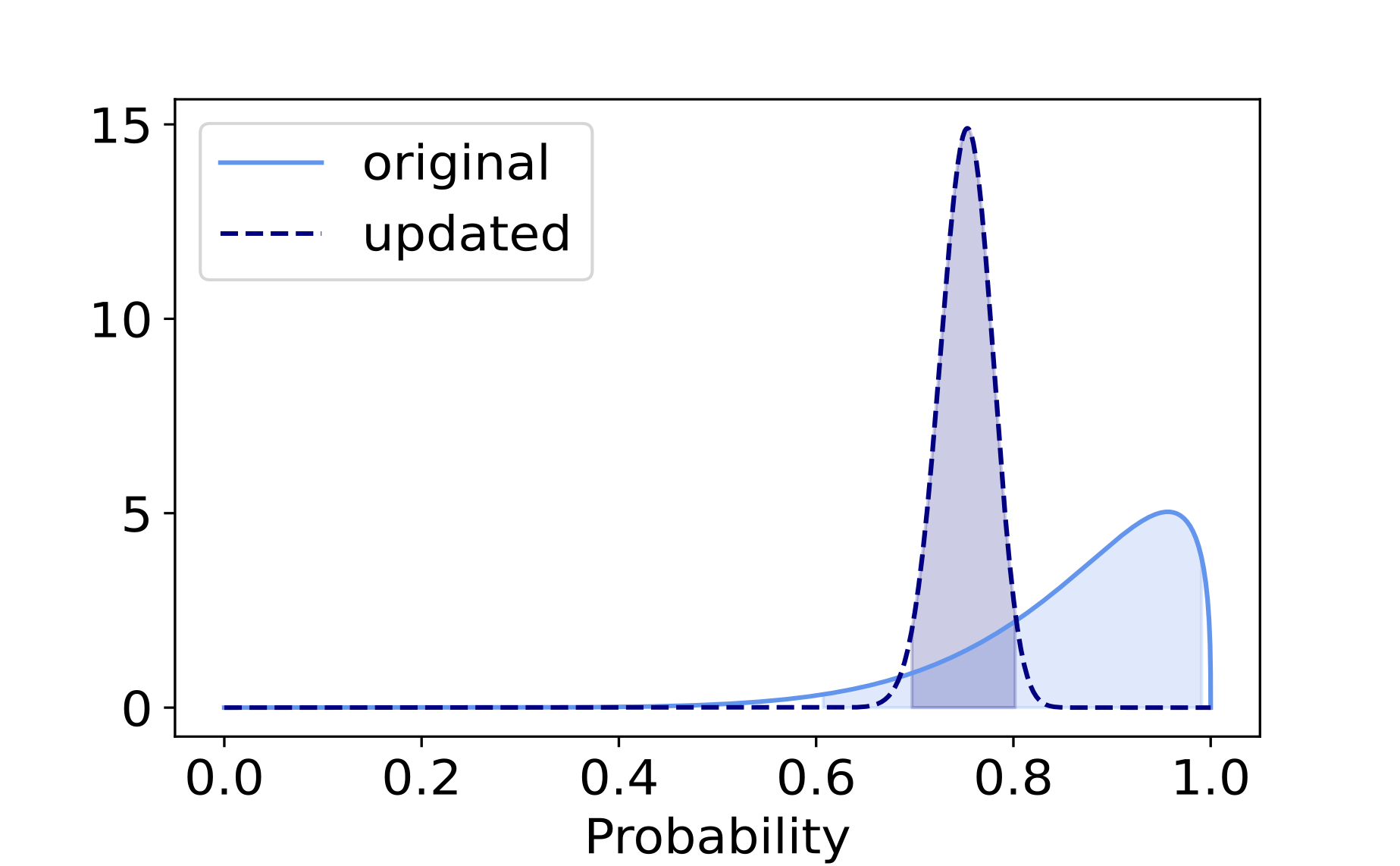}} &
        \subfloat[]{\includegraphics[width=0.28\textwidth]{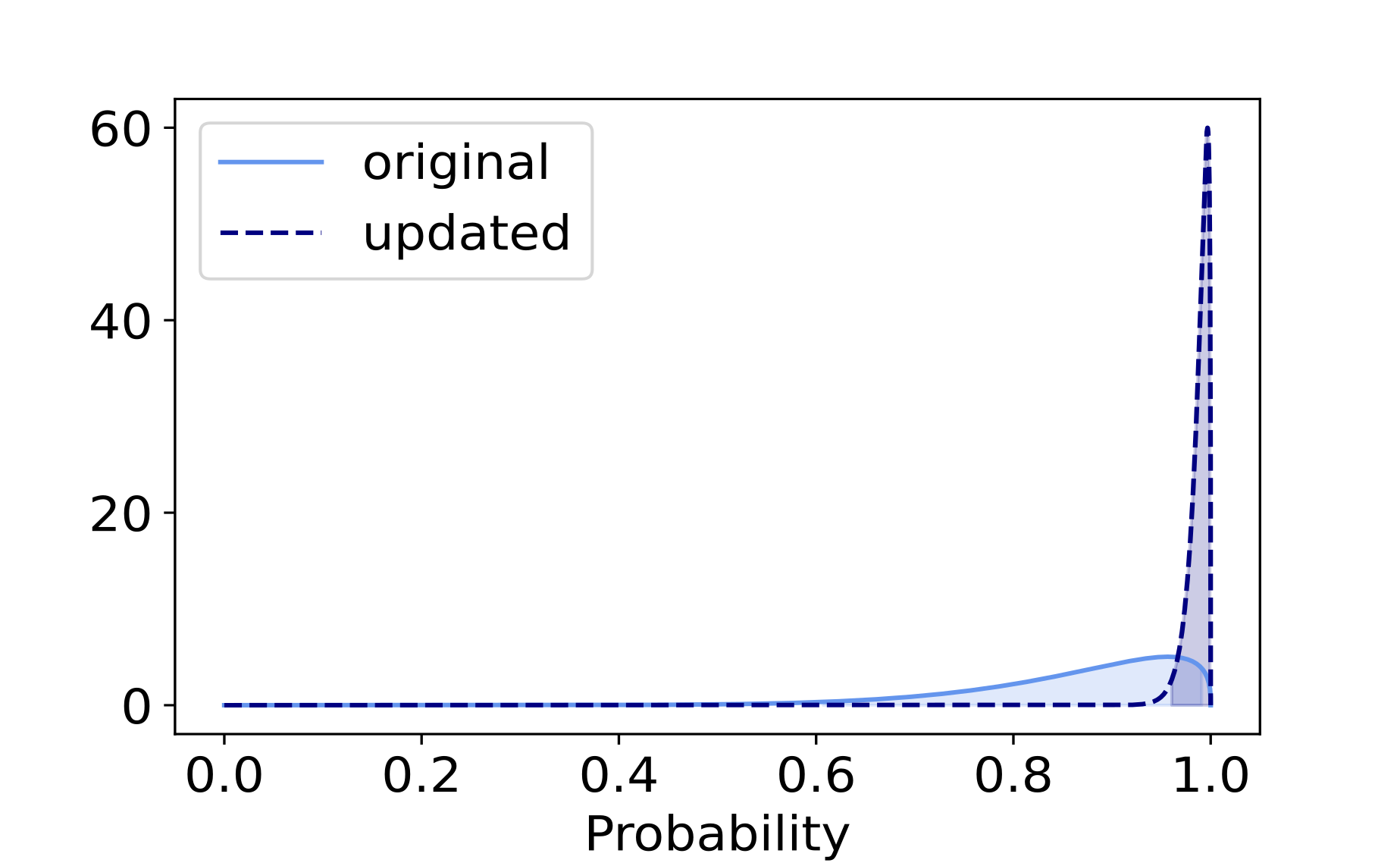}} &
        \subfloat[]{\includegraphics[width=0.28\textwidth]{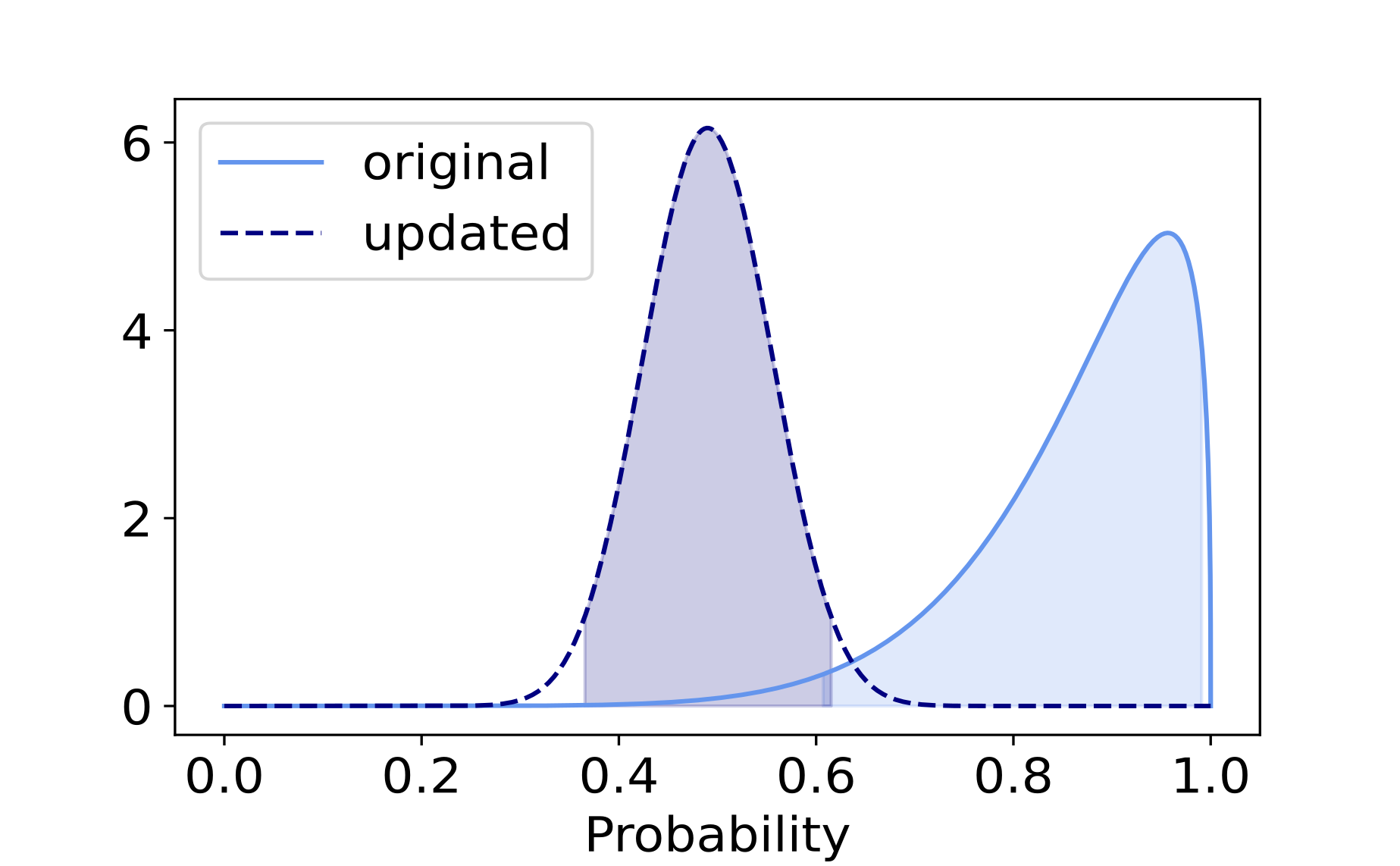}} \\
    \end{tabular}
    \caption{Three scenarios demonstrating the effect of SPI monitoring and SL-based update on the resulting claim opinion.}
    \label{fig:additional_experiments}
\end{figure*}

The impact of the SPI on the claim is further illustrated across several scenarios in Fig.~\ref{fig:additional_experiments}. In the first scenario (a), an initially uncertain opinion with high belief shifts to a lower-belief opinion with reduced uncertainty. Here, the uncertainty mass shifts toward disbelief as SPI violations accumulate, providing evidence against the claim. In the second scenario (b), we observe a segment without SPI violations. As more supporting evidence is gathered, uncertainty gradually decreases, and belief strengthens. In both cases, uncertainty is reduced over time, with the mass distributed toward either belief or disbelief based on the nature of the evidence collected at runtime. The final scenario (c) presents a shorter nighttime segment, where some uncertainty remains due to insufficient evidence being gathered during runtime. This indicates that further data collection is required to form a more confident opinion.

\section{Related work}\label{sec:related_work}
The idea of \textit{dynamic safety cases} was first presented by Denney \textit{et al.} \cite{denney_2015}. The authors propose a general framework with basic principles and lifecycle activities to achieve \textit{through-life assurance}. The framework is useful in framing our work in this paper: first, we address two of the stated principles -- \textit{(1) proactively computing the confidence in, and update the reasoning about, the safety of ongoing operations} and \textit{(2) providing an increased level of formality in the safety infrastructure}. Second, our work addresses two of the four mentioned lifecycle stages in particular: \textit{(1) identify} the sources of uncertainty, and \textit{(3) analyze} the operational data and update the confidence in the associated claims.    

In more recent work, Denney \textit{et al.} \cite{denney2024reconciling} present a dynamic safety case approach grounded in classical probability: runtime counts update Bayesian priors, and revised probabilities are propagated through the safety architecture to reassess operational risk and detect drift. Their indicators are tightly linked to safety case elements and support systematic, architecture-level risk management. In contrast, our approach is claim-centric within a single SL-based argument, makes (first-order and second-order) uncertainty explicit, and uses an immediate-refutation rule on windowed leading SPIs to prioritize safety responsiveness. We view the approaches as complementary: architecture-level risk quantification from \cite{denney2024reconciling} can be combined with our opinion-based claim confidence to provide both quantitative risk propagation and responsive, auditable governance.

Schleiss et al. \cite{Schleiss2022} propose a unified continuous assurance framework which aligns terminology from existing work. They focus on safety as an absence of unreasonable risk, and highlight the need for a systematic quantification of uncertainty to address this risk. Within the framework, different categories of monitors are proposed based on their target uncertainty: known-(un)known, unknown-unknown, and ACP monitoring, where the latter two are considered assurance uncertainty monitors. The SPI monitors proposed in our work align with this framework and fit into the category of assurance uncertainty monitoring. 

Weyns et al. \cite{weyns2017perpetual} provide another framework for \textit{perpetual assurance}, i.e.\ the ongoing integration of evidence into the assurance process for self-adaptive systems. They put a particular emphasis on different sources and types of uncertainty and derive a set of requirements that a perpetual assurance process should satisfy along with benchmark criteria. Whilst the authors focus more on the overall building blocks required for perpetual assurance, our work can be seen as providing one concrete method for satisfying some of these requirements such as continually observing, quantifying, and reducing the sources of uncertainty.    

Hawkins and Ryan Conmy \cite{hawkinsIdentifyingRunTimeMonitoring2023} use a dialectic approach to systematically analyse a safety case and determine monitoring requirements. The approach is based on constructing an \textit{operational dialectic argument}, whereby potential runtime \textit{challenges} are identified and linked to the respective claims which they question. Each challenge is supported by counter-evidence, which can be monitored at runtime to identify potential claim violations based on associated triggers. In our work, we quantitatively evaluate runtime evidence to update confidence in a claim, providing additional insights into the (positive or negative) impact of counter-evidence on overall argument confidence.
. 

Ratiu et al. \cite{ratiu2024towards} present an argument structure about the use of SPIs to maintain the validity of a safety case. The structure contains subgoals referring to the sufficiency of the defined SPIs, the collection and analysis of SPI violations, and the responses to SPI violations. To establish confidence in the SPIs themselves, the authors further introduce a set of \textit{meta-SPIs} in the form of monitors that act as defeaters and are integrated as challenge claims into the argument. The approach assumes a qualitative argument where each SPI violation refutes associated safety claim, but could also be adapted to a quantitative argument. 

\section{Discussion and future work}\label{sec:conclusions}

This paper presents a method for integrating Safety Performance Indicators (SPIs) into Subjective Logic (SL)-based quantitative assurance arguments for the purpose of dynamic safety assurance in evolving operational environments. By dynamically incorporating runtime information through SPIs, the approach addresses the challenge of maintaining the validity of assurance arguments by continuously updating confidence in safety claims in a quantitative and automated way. The combination of two SL operators (cumulative fusion and refuting challenger) provides a robust formal mechanism for confidence updating, allowing for the representation of both the accumulation of positive evidence and the immediate negative impact of SPI violations, as illustrated in the examples.

With this work, we hope to contribute to an increased formalization of continuous assurance for AI-based and autonomous systems that must operate reliably in complex, changing environments. By aligning with standards like UL 4600, we offer a practical approach for ongoing validation of safety claims and the deployment of functions and systems in evolving environments with greater assurance.

Dynamic updates to an assurance argument serve different stakeholders with distinct needs. For example, fleet operators and safety managers may use SPI-driven claim updates during runtime to monitor trends, trigger mitigations such as parameter adjustments, scenario blacklisting, or additional human oversight, and to enforce decision policies (for example, escalating if disbelief exceeds a certain threshold). This provides a governance layer in the safety management system, which is especially important for ML-based functions operating in open, evolving contexts. Regulators and auditors may prefer periodic summaries and audit trails over live streams of runtime signals to review how runtime evidence affects specific claims, how violations were handled, and how operational decisions were justified; the SL-based argument thus offers an auditable link from runtime signals to the safety case. Engineering teams can apply the same mechanism in simulation and pilot deployments but also during testing or in shadow mode to gate data collection and trigger targeted retraining or fine-tuning when assurance confidence degrades. This creates a link between assurance and MLOps by connecting opinion thresholds with actions such as focused data acquisition, labeling, retraining, and staged deployment. To avoid unstable back-and-forth reactions, triggers can be set that act only when a drop in confidence persists for a defined period or number of observations (e.g., by setting a `minimum evidence window'), and record all decisions for audit. The proposed approach thus adds an additional layer on top of raw SPI signal streams. The SL layer maps signals to explicit claim-level opinions with first-order and second-order confidence, allows for the composition of effects across the argument via SL, and enables the targeted, non-Bayesian refutation of claims when violations occur. This yields a transparent confidence narrative that supports governance and continuous oversight.

The approach is, of course, not without limitations. The effectiveness depends heavily on accurate runtime data collection and the appropriate setting of SPI thresholds, which may require extensive tuning and validation in real-world applications. Additionally, computational requirements for real-time SPI monitoring and confidence updating could pose challenges for integration into existing systems. Future research should focus on refining the parameters used in SPI assessments and exploring alternative SL operators to enhance the flexibility and accuracy of confidence updates. One step towards practical applicability could be to investigate existing SPI libraries and formalize them using the proposed approach. This paper currently focuses on \textit{windowed runtime SPIs} and their integration into SL updates with immediate refutation. However, several SPIs are quantitative yet update infrequently. One practical way to treat them is to assess them at design time and at scheduled operational reassessments, updating the associated opinion via cumulative fusion, and to apply immediate refutation only when a periodic check fails. To mitigate staleness, time-based discounting can be used to increase uncertainty ($u$) in the absence of fresh evidence, or to attach leading runtime drift monitors as proxies to trigger reassessment of the static SPI. Safety arguments often need both early-warning \emph{leading} SPIs and slower, outcome-based \emph{lagging} SPIs; to avoid double counting and mismatched update frequencies, they should be modeled as separate subclaims that feed a parent claim: leading SPIs act as immediate challengers imposing targeted penalties on violations, while lagging SPIs update via cumulative fusion at reassessments, increasing precision as sufficient outcome data accrue. Priors and contexts should be aligned before combination. Formalizing these composition patterns, correlation handling, and discounting policies is left to future work. Finally, we aim to relax the current assumption of independent, non-overlapping SPI windows by explicitly modeling temporal correlation between windows and analyzing its effect on confidence updates.

Overall, integrating SPIs into quantitative assurance arguments is an important step toward a more formal treatment of dynamic safety cases. Our SL-based update method supports a responsive approach to safety validation and helps maintain confidence in safety claims as operational conditions evolve. We hope this work advances more resilient and adaptable methods for continuous assurance to support safer deployment in safety-critical contexts.

\begin{acks}
The research leading to these results is funded by the German Federal Ministry for Economic Affairs and Energy within the project ``Safe AI Engineering –- Sicherheitsargumentation befähigendes AI Engineering über den gesamten Lebenszyklus einer KI-Funktion''. The authors would like to thank the consortium for the successful cooperation and the reviewers for their helpful comments. ChatGPT was utilized to improve wording and to spell-check this paper. 
\end{acks}

\printbibliography

@inproceedings{denney_2015,
  author={Ewen Denney and Ganesh Pai and Ibrahim Habli},
  booktitle={{2015 IEEE/ACM 37th IEEE International Conference on Software Engineering}}, 
  title={Dynamic Safety Cases for Through-Life Safety Assurance}, 
  year={2015},
  volume={2},
  pages={587-590},
  doi={10.1109/ICSE.2015.199}}

@inproceedings{Schleiss2022,
  author       = {Philipp Schleiss and
                  Francesco Carella and
                  Iwo Kurzidem},
  title        = {Towards Continuous Safety Assurance for Autonomous Systems},
  booktitle    = {{ICSRS}},
  pages        = {457--462},
  publisher    = {{IEEE}},
  year         = {2022}
}

@book{josang2016subjective,
  title={{Subjective Logic}},
  author={J{\o}sang, Audun},
  volume={3},
  year={2016},
  publisher={Springer}
}

@misc{UL4600_2023,
  title        = "{UL 4600: Evaluation of Autonomous Products}",
  author       = "{ANSI/UL 4600}",
  year         = {2023},
  edition      = {3rd},
  publisher    = "{Underwriters’ Laboratories}",
  address      = "{Northbrook, IL}",
}

@article{Wang2019,
title = {Safety case confidence propagation based on {Dempster–Shafer theory}},
journal = {Int. Journal of Approx. Reasoning},
volume = {107},
pages = {46-64},
year = {2019},
issn = {0888-613X},
doi = {https://doi.org/10.1016/j.ijar.2019.02.002},
author = {Rui Wang and Jérémie Guiochet and Gilles Motet and Walter Schön}
}

@INPROCEEDINGS{Guo2003,
  author={Guo, B.},
  booktitle={International Conference on Natural Language Processing and Knowledge Engineering, 2003. Proceedings. 2003}, 
  title={Knowledge representation and uncertainty management: applying {Bayesian Belief Networks} to a safety assessment expert system}, 
  year={2003},
  volume={},
  number={},
  pages={114-119},
  doi={10.1109/NLPKE.2003.1275879}}

@InProceedings{Hobbs2012,
author="Hobbs, Chris
and Lloyd, Martin",
editor="Dale, Chris
and Anderson, Tom",
title="The Application of {Bayesian Belief Networks} to Assurance Case Preparation",
booktitle="Achieving Systems Safety",
year="2012",
publisher="Springer London",
address="London",
pages="159--176",
isbn="978-1-4471-2494-8"
}

@inproceedings{duan2016representation,
  title={Representation of confidence in assurance cases using the beta distribution},
  author={Duan, Lian and Rayadurgam, Sanjai and Heimdahl, Mats and Sokolsky, Oleg and Lee, Insup},
  booktitle={2016 IEEE 17th International Symposium on High Assurance Systems Engineering (HASE)},
  pages={86--93},
  year={2016},
  organization={IEEE}
}

@INPROCEEDINGS{Goodenough2013,
  author={Goodenough, John B. and Weinstock, Charles B. and Klein, Ari Z.},
  booktitle={2013 35th Int. Conf. on Software Engineering (ICSE)}, 
  title={Eliminative induction: A basis for arguing system confidence}, 
  year={2013},
  volume={},
  number={},
  pages={1161-1164},
  doi={10.1109/ICSE.2013.6606668}
}

@inproceedings{Ayoub2013,
  title={Assessing the overall sufficiency of safety arguments},
  author={Ayoub, Anaheed and Chang, Jian and Sokolsky, Oleg and Lee, Insup},
  booktitle={21st Safety-critical Systems Symposium (SSS’13), Bristol, United Kingdom},
  pages={127-144},
  year={2013}
}

@article{yuan2017subjective,
  title={A subjective logic-based approach for assessing confidence in assurance case},
  author={Yuan, Chunchun and Wu, Ji and Liu, Chao and Yang, Haiyan},
  journal={International Journal of Performability Engineering},
  volume={13},
  number={6},
  pages={807},
  year={2017}
}

@INPROCEEDINGS{Denney2011,
  author={Denney, Ewen and Pai, Ganesh and Habli, Ibrahim},
  booktitle={2011 International Symposium on Empirical Software Engineering and Measurement}, 
  title={Towards Measurement of Confidence in Safety Cases}, 
  year={2011},
  volume={},
  number={},
  pages={380-383},
  doi={10.1109/ESEM.2011.53}
}

@ARTICLE {herd2024,
    author  = "Herd, Benjamin and Burton, Simon",
    title   = "{Can you trust your ML metrics? Using Subjective Logic to determine the true contribution of ML metrics for safety}",
    journal = "Proc. of the 39th ACM/SIGAPP Symposium On Applied Computing (SAC24)",
    year    = "2024"
}

@inproceedings{hawkinsIdentifyingRunTimeMonitoring2023,
	location = {Cham},
	title = {Identifying Run-Time Monitoring Requirements for Autonomous Systems Through the Analysis of Safety Arguments},
	isbn = {978-3-031-40923-3},
	doi = {10.1007/978-3-031-40923-3_2},
	pages = {11--24},
	booktitle = {Computer Safety, Reliability, and Security},
	publisher = {Springer Nature Switzerland},
	author = {Hawkins, Richard and Ryan Conmy, Philippa},
	editor = {Guiochet, Jérémie and Tonetta, Stefano and Bitsch, Friedemann},
	date = {2023},
	langid = {english},
}

@ARTICLE {herd2025,
    author  = "Herd, Benjamin and Kelly, Jessica and Heinemann, Clarissa and Zacchi, Jo{\~a}o-Vitor",
    title   = "{Integrating Defeaters into Subjective Logic-based Quantitative Assurance Arguments}",
    journal = "Proc. of the 20th European Dependable Computing Conf. (EDCC)",
    year    = "2025",
}

@inproceedings{ratiu2024towards,
  title={Towards an Argument Pattern for the Use of Safety Performance Indicators},
  author={Ratiu, Daniel and Rohlinger, Tihomir and Stolte, Torben and Wagner, Stefan},
  booktitle={Int. Conference on Computer Safety, Reliability, and Security},
  pages={160--172},
  year={2024},
  organization={Springer}
}

@techreport{graydonDefiningBaconianProbability2016,
  title = {Defining {{Baconian Probability}} for {{Use}} in {{Assurance Argumentation}}},
  author = {Graydon, Patrick J.},
  date = {2016-10-01},
  institution = {NASA Langley Research Center},
  year = {2016},
  number = {NASA/TM-2016-219341},
}

@inproceedings{weyns2017perpetual,
  title={Perpetual assurances for self-adaptive systems},
  author={Weyns, Danny and Bencomo, Nelly and Calinescu, Radu and Camara, Javier and Ghezzi, Carlo and Grassi, Vincenzo and Grunske, Lars and Inverardi, Paola and Jezequel, Jean-Marc and Malek, Sam and others},
  booktitle={Software Engineering for Self-Adaptive Systems III. Assurances: International Seminar, Dagstuhl Castle, Germany, December 15-19, 2013, Revised Selected and Invited Papers},
  pages={31--63},
  year={2017},
  organization={Springer}
}

@techreport{IEEE_15026,
  author =          "{ISO}",
  title =          "Systems and software engineering: Systems and software assurance ",
  number =          "ISO/IEC/IEEE 15026:2019",
  institution =  "{Int. Organization for Standardization}",
  year =          2019
}

@misc{yolov8_ultralytics,
  author = {Glenn Jocher and Ayush Chaurasia and Jing Qiu},
  title = {{Ultralytics YOLOv8}},
  version = {8.0.0},
  year = {2023},
  url = {https://github.com/ultralytics/ultralytics},
}

@article{ros2,
    author = {Steven Macenski  and Tully Foote  and Brian Gerkey  and Chris Lalancette  and William Woodall },
    title = {Robot Operating System 2: Design, architecture, and uses in the wild},
    journal = {Science Robotics},
    volume = {7},
    number = {66},
    pages = {eabm6074},
    year = {2022},
}

@inproceedings{carla,
  author       = {Alexey Dosovitskiy and
                  Germ{\'{a}}n Ros and
                  Felipe Codevilla and
                  Antonio M. L{\'{o}}pez and
                  Vladlen Koltun},
  title        = {{CARLA:} An Open Urban Driving Simulator},
  booktitle    = {CoRL},
  series       = {Proceedings of Machine Learning Research},
  volume       = {78},
  pages        = {1--16},
  publisher    = {{PMLR}},
  year         = {2017}
}

@inproceedings{denney2024reconciling,
  title={Reconciling Safety Measurement and Dynamic Assurance},
  author={Denney, Ewen and Pai, Ganesh},
  booktitle={Int. Conf. on Computer Safety, Reliability, and Security},
  pages={51--67},
  year={2024},
  organization={Springer}
}

\end{document}